\documentclass[pdflatex,sn-mathphys-num]{sn-jnl}% Math and Physical Sciences Numbered Reference Style 
\usepackage{graphicx}%
\usepackage{multirow}%
\usepackage{amsmath,amssymb,amsfonts}%
\usepackage{amsthm}%
\usepackage{mathrsfs}%
\usepackage[title]{appendix}%
\usepackage{xcolor}%
\usepackage{textcomp}%
\usepackage{manyfoot}%
\usepackage{booktabs}%
\usepackage{algorithm}%
\usepackage{algorithmicx}%
\usepackage{algpseudocode}%
\usepackage{listings}%
\usepackage{pifont}% http://ctan.org/pkg/pifont
\newcommand{\cmark}{\ding{51}}%
\newcommand{\xmark}{\ding{55}}%

\begin{document}

\title[Article Title]{IGCN: Integrative Graph Convolution Networks for patient level insights and biomarker discovery in multi-omics integration}


\author[1,3,4]{\fnm{{Cagri}} \sur{Ozdemir$^{\text{1,3,4}}$}}\email{cagri.ozdemir@unt.edu}

\author[1,3,4]{\fnm{Yashu} \sur{Vashishath$^{\text{1,3,4}}$}}\email{yashu.vashishath@unt.edu}

\author[1,2,3,4]{\fnm{Serdar} \sur{Bozdag$^{\text{1,2,3,4}}$}}\email{serdar.bozdag@unt.edu}

\author{and for the Alzheimer’s Disease Neuroimaging Initiative}
\equalcont{\scriptsize{Data used in preparation of this article were obtained from the Alzheimer’s Disease
Neuroimaging Initiative (ADNI) database (adni.loni.usc.edu). As such, the investigators within the ADNI contributed to the design and implementation of ADNI and/or provided data but did not participate in analysis or writing of this report. A complete listing of ADNI investigators can be found at: \url{https://adni.loni.usc.edu/wp-content/uploads/how_to_apply/ADNI_Acknowledgement_List.pdf}}}

\affil[1]{\normalsize{\orgdiv{Department of Computer Science and Engineering}, \orgname{University of North Texas}, \orgaddress{\city{Denton}, \state{TX}, \country{USA}}}}

\affil[2]{\normalsize{\orgdiv{Department of Mathematics}, \orgname{University of North Texas}, \orgaddress{\city{Denton}, \state{TX}, \country{USA}}}}

\affil[3]{\normalsize{\orgdiv{BioDiscovery Institute}, \orgname{University of North Texas}, \orgaddress{\city{Denton}, \state{TX}, \country{USA}}}}

\affil[4]{\normalsize{\orgdiv{Center for Computational Life Sciences USA}, \orgname{University of North Texas}, \orgaddress{\city{Denton}, \state{TX}, \country{USA}}}}


\abstract{Developing computational tools for integrative analysis across multiple types of omics data has been of immense importance in cancer molecular biology and precision medicine research. While recent advancements have yielded integrative prediction solutions for multi-omics data, these methods lack a comprehensive and cohesive understanding of the rationale behind their specific predictions. To shed light on personalized medicine and unravel previously unknown characteristics within integrative analysis of multi-omics data, we introduce a novel integrative neural network approach for cancer molecular subtype and biomedical classification applications, named Integrative Graph Convolutional Networks (IGCN). IGCN can identify which types of omics receive more emphasis for each patient to predict a certain class. Additionally, IGCN  has the capability to pinpoint significant biomarkers from a range of omics data types. To demonstrate the superiority of IGCN, we compare its performance with other state-of-the-art approaches across different cancer subtype and biomedical classification tasks.}

\maketitle

\section{Introduction}\label{sec:sec1}
Due to advancements in biotechnology, innovative omics technologies are constantly emerging, allowing researchers to access multi-layered information from genome-wide data. These multi-layered information can be obtained for the same set of samples, leading to generation of multi-omics datasets. Cancer molecular subtype prediction is a crucial area of research focused on classifying cancers into distinct subtypes based on their molecular characteristics. These subtypes can provide valuable insights into the biology of the cancer, predict patient outcomes, and guide personalized treatment strategies~\cite{chaudhary2018deep,poirion2018deep}. Predicting cancer molecular subtypes through multi-omics integration may reveal complex interactions within biological systems and shed light on molecular mechanisms that contribute to cancer development and progression, which might be missed when examining a single type of omics data alone~\cite{sharifi2019moli,huang2019salmon,choi2023mobrca,khadirnaikar2023machine,gong2023multi}. Moreover, a multitude of approaches have shown that integration of multi-omics data can contribute to the precision medicine efforts in diseases such as Alzheimer's Disease (AD).~\cite{abbas2022alzheimer,shigemizu2020prognosis,li2021applied}.
% ~\cite{shen2009integrative,wang2014similarity,huang2019salmon,singh2019diablo,he2023artificial}
While deep neural networks (NN)-based methods have been introduced as multi-omics integrative tools for cancer subtype prediction and  diverse biomedical classification tasks~\cite{chaudhary2018deep,poirion2018deep,sharifi2019moli,huang2019salmon,choi2023mobrca,khadirnaikar2023machine,gong2023multi}, graph neural network (GNN)-based multi-omics integration approaches have shown promising results~\cite{wang2021mogonet,kesimoglu2022supreme,yin2022molecular,xiao2023graph}.

Multi-omics graph convolutional networks (MOGONET) has been introduced as a supervised multi-omics integration framework for a wide range of
biomedical classification applications, which uses separate graph convolutional networks (GCN) for patient similarity networks based on mRNA expression, DNA methylation, and microRNA (miRNA) expression data types~\citep{wang2021mogonet}.
% Comparing with the fully connected neural networks,
% in MOGONET, GCN layers leverage both omics features and  correlations among samples described by similarity networks to enhance classification performance.
MOGONET also utilizes View Correlation Discovery Network (VCDN) to explore the cross-omics correlations at the label space for effective multi-omics integration. Another computational tool named SUPREME, a subtype prediction methodology, integrates multiple types of omics data using GCN~\citep{kesimoglu2022supreme}. To obtain embeddings, SUPREME concatenates the features from all omics data types in the input space and utilizes them as node attributes in each GCN module to derive embeddings. Subsequently, SUPREME integrates these embeddings and conducts comprehensive evaluations of all possible combinations. Most recently, Trusted Multi-Omics integration framework based on hypergraph convolutional networks (called HyperTMO) has been developed~\cite{wang2024hypertmo}. HyperTMO constructs hypergraph structures to represent the associations between samples in single-omics data. Following that, feature extraction is conducted utilizing a hypergraph convolutional network, while multi-omics integration occurs during the late stages of analysis.

Besides these GNN-based integrative tools applied for multi-omics datasets, several GNN-based approaches have been introduced  as more general integrative computational tools for multi-modal datasets. Relational Graph Convolutional Networks (RGCN)~\citep{schlichtkrull2018modeling} provides relation-specific transformations, i.e. depending on the type and direction of an edge, for
large-scale and multi-modal data. Heterogeneous Graph Attention Network (HAN)~\citep{wang2019heterogeneous} generates meta-path-based networks from a multi-modal network. The concept of meta-path can be applied to learn 
a sequence of relations defined between different objects in a multi-modal graph~\citep{sun2011pathsim}. After generating meta-paths, HAN takes node-level attention (for each node using its meta-path-based neighborhood) and association-level attention (for each meta-path) into consideration simultaneously. HAN employs a multi-layer perceptron (MLP) module to compute class probabilities rather than taking advantage of graph topologies, which could potentially limit the ability of the model to capture complementary information from graph topologies to make predictions. 
% Furthermore, it has been argued that vanilla GCN and GAT methods could outperform HAN after preparing the network in a more fair way~\citep{lv2021we}.
Apart from HAN, in the context of handling graph heterogeneity, Heterogeneous Graph Transformer (HGT)~\citep{hu2020heterogeneous} has been developed to maintain representations dependent on node and edge types.

As outlined in \cite{lv2021we}, it has been argued that vanilla GCN and GAT methods could outperform existing integrative approaches after making some modifications to the networks. This underscores the necessity for more advanced computational methodologies in the integrative analysis of multi-omics data. In addition, the existing tools have some limitations. Omics data, by their nature, do not inherently exhibit a graph structure, thus a graph construction procedure is needed. However, constructing a graph from omics data could suffer from data noise due to many possible factors such as measurement inaccuracies, missing values, or inherent fluctuations within the dataset. These factors can adversely affect tasks like clustering, classification, or link prediction. Additionally, in the context of biomedical classification, different types of omics data have the potential to reveal unique characteristics at the label space. In other words, some omics types may demonstrate superior performance when predicting one disease label, while others might excel in predicting a different disease type. Therefore, directly fusing different types of omics data without considering the sample level importance of different omics networks may be liable to make wrong predictions and cause some level of performance degradation. Furthermore, multitudes of existing multi-omics integration tools do not explain how and why their models came to the prediction. In predictive modeling, a crucial trade-off arises: Do we merely desire the prediction, or are we interested in understanding the rationale behind it?~\citep{doshi2017towards}. Since each of multi-omics type captures a different part of the underlying biology, understanding the 'why' can contribute to a deeper comprehension of the problem and advance the road toward precision medicine.

To address these limitations, we introduce a novel supervised integrative graph convolutional networks (IGCN) architecture that operates on multi-omics data structures. In IGCN, a multi-GCN module is initially employed to extract node embeddings from each network. A personalized attention module is then proposed to fuse the multiple node embeddings into a weighted form. Unlike previous multi-omics integration studies, the attention mechanism assigns different attention coefficients to each node/sample for each data modality to help identify which data modality receives more emphasis to predict a certain class type. This feature makes IGCN interpretable in terms of understanding the rationale behind the prediction at the sample level. Furthermore, IGCN has the capability to assign attention coefficients to features for each sample from a range of omics data types, which would facilitate identifying omics biomakers associated with phenotypes of interest. To the best of our knowledge, IGCN stands out as the first supervised integrative approach that provides patient level insights and biomarkers in multi-omics integration. 

We presented our experimental results on four classification tasks: breast invasive carcinoma and glioblastoma (GBM) molecular subtype classification using The Cancer Genome Atlas dataset~\cite{colaprico2016tcgabiolinks}; Alzheimer’s
Disease (AD) patients vs. cognitively normal (CN)
classification using  The Religious Orders Study and Memory and Aging Project (ROSMAP) cohort~\cite{hodes2016accelerating}; and AD, Mild Cognitive Impairment (MCI), and CN classification task using Alzheimer’s Disease Neuroimaging Initiative (ADNI) dataset~\cite{petersen2010alzheimer}. Our experimental results show that our
proposed model outperforms the state-of-the-art and baseline methods. IGCN identifies which types of omics data receive more emphasis for each patient when predicting a specific class. Additionally, IGCN has the capability to pinpoint significant biomarkers from a range of omics data types.

\section{Results}
\label{sec:sec2}
\subsection{Overview of IGCN architecture}
\begin{figure*}[b!]
    \centering
    \includegraphics[width=1\textwidth]{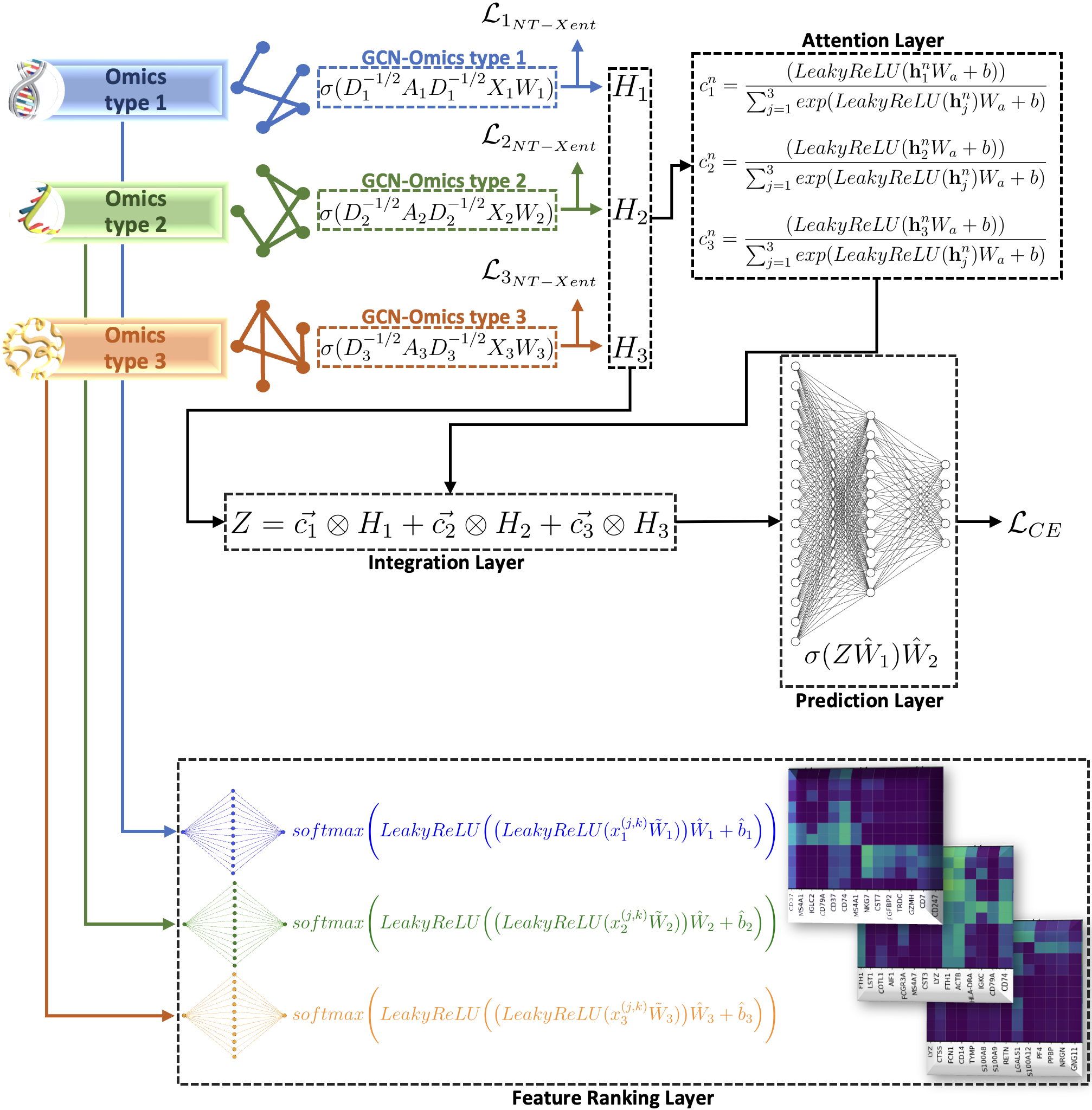}
    \caption{\textbf{Overview of IGCN architecture on three similarity networks.} IGCN employs an integration module to fuse the node embeddings. Simultaneously, IGCN assigns attention coefficients to features for individual samples across diverse omics data types.}
    \label{fig:IGCN}
\end{figure*}

IGCN integrates multi-omics data and reveals patient level insights regarding both the key omics types and features for biomedical classification tasks. The overview of IGCN architecture is illustrated in \textbf{Fig.~(\ref{fig:IGCN}}). We first construct graphs for each omics type (Eq.~(\ref{eq:adj_mtx}) and Eq.~(\ref{eq:treshold})). Subsequently, IGCN utilizes GCN modules on graphs to learn the node embeddings. As depicted in Algorithm~\ref{<alg-label>}, the hyperparameter $\epsilon$ determines a threshold for correlation in graph construction. In graph construction, it is common to encounter data noise from various sources, such as measurement inaccuracies, missing values, or inherent dataset fluctuations. To alleviate this noise, the Normalized Temperature-scaled Cross Entropy loss (NT-Xent loss) (Eq.~(\ref{eq:NT-xent})) is utilized for each GCN module. Following this, we introduce a personalized attention module to merge the multiple node embeddings into a weighted representation (Eq.~(\ref{eq:at_coef}) and Eq.~(\ref{eq:w_form})). Differing from prior research, this attention mechanism assigns distinct attention coefficients to each patient, facilitating the identification which data modality is more influential in predicting a particular class type. Similar to the attention mechanism module, the feature ranking module also offers personalized insights regarding
feature importance (Eq.~(\ref{eq:feature_att})). To our knowledge, IGCN is the first supervised integrative approach that offers patient-level insights and biomarkers in multi-omics integration, making it a significant advancement in the field.

\subsection{Biomedical classification tasks}
To perform breast invasive carcinoma (BRCA) subtype classification task, mRNA expression, DNA methylation, and miRNA expression data were collected from TCGA project data portal. PAM50 labels of the tumor samples were obtained as the ground truth class labels~\citep{parker2009supervised}. Specifically, we have five different class labels: Basal-like, HER2-Enriched, Luminal-A, Luminal-B, and Normal-like. For GBM molecular subtype classification task, we acquired mRNA expression and miRNA expression data from the the Broad Institute's Firehose data portal~\cite{deng2017firebrowser} (available at \url{http://firebrowse.org/}). DNA methylation data was not used because most of the samples available for other data modalities were not available for DNA methylation, leading to a small sample size. We utilized the ground truth labels provided in~\cite{verhaak2010integrated}, which identify four molecular subtypes based on gene expression, namely Proneural, Neural, Mesenchymal, and Classical.

In addition to the cancer molecular subtype classification tasks, we also conducted biomedical classification analyses for AD. mRNA expression, DNA methylation, and miRNA expression data were collected from ROSMAP cohort~\cite{hodes2016accelerating} to classify between AD and CN individuals. We also collected diverse omics types, namely single
nucleotide polymorphisms (SNPs), lipidomics, and bileomics, from ADNI dataset~\cite{petersen2010alzheimer} and conduct a classification task to predict AD, MCI, and CN individuals. A detailed description of the datasets can be found in \textbf{Table S1}.

\subsection{IGCN demonstrated superior performance in various cancer molecular subtype prediction and biomedical classification tasks}
We compared the performance of IGCN with the state-of-the-art methods (i.e., GCN, GAT, HAN, HGT, RGCN, MOGONET, SUPREME, and HyperTMO) as well as baseline methods, namely MLP, Random Forest (RF), and Support Vector Machine (SVM). We evaluated their performance based on four metrics: accuracy, macro F1 score, weighted F1 score, and Matthew's correlation coefficient (MCC). We evaluated all the methods on ten different randomly generated training, validation, and test splits. We selected  80\% of the samples as the training set and 20\% of the samples as the test set in a stratified fashion. We also used 25\% of the training set as the validation set to tune the hyperparameters (e.g., hidden layer size and learning rate) and perform early stopping.

% Since GCN, GAT, MLP, SVM, and RF are not integrative tools, we evaluated each data modality separately and presented the best results for each method. We also note that SUPREME can be computed on different combinations of the similarity networks. For each run, we selected the best combination that gives the best macro F1 score on the validation data. 

The quantitative results of our comparative experiments, presented in \textbf{Table~\ref{tab:ex_res}}, show that IGCN achieved the best performance across all metrics and datasets. Since GCN, GAT, MLP, SVM, and RF are not integrative tools, we evaluated each data modality separately and presented the best results for each method. 
% For these tools, the best performance was observed with the mRNA expression data on 
% breast and GBM subtype classification tasks, and with the SNP data on AD classification task. 
These results highlight the importance of multi-omics integration because even the top performance achieved is still not superior to that of IGCN.

In \cite{lv2021we}, it was shown that given proper inputs, simple homogeneous GNN-based integration approaches, such as GCN and GAT, may surpass the performance of all existing integrative tools across various scenarios. Similarly, our experimental results show that GCN delivers the second-best performance on GBM dataset, outperforming integrative methods such as HAN, HGT, RGCN, MOGONET, SUPREME, and HyperTMO. Furthermore, both GCN and GAT outperformed both MOGONET and HGT on ADNI dataset. However, IGCN demonstrated the best classification performance across all dataset, representing a more sophisticated and advanced integrative tool.

\begin{figure}[h!]
    \centering
    \includegraphics[width=\textwidth]{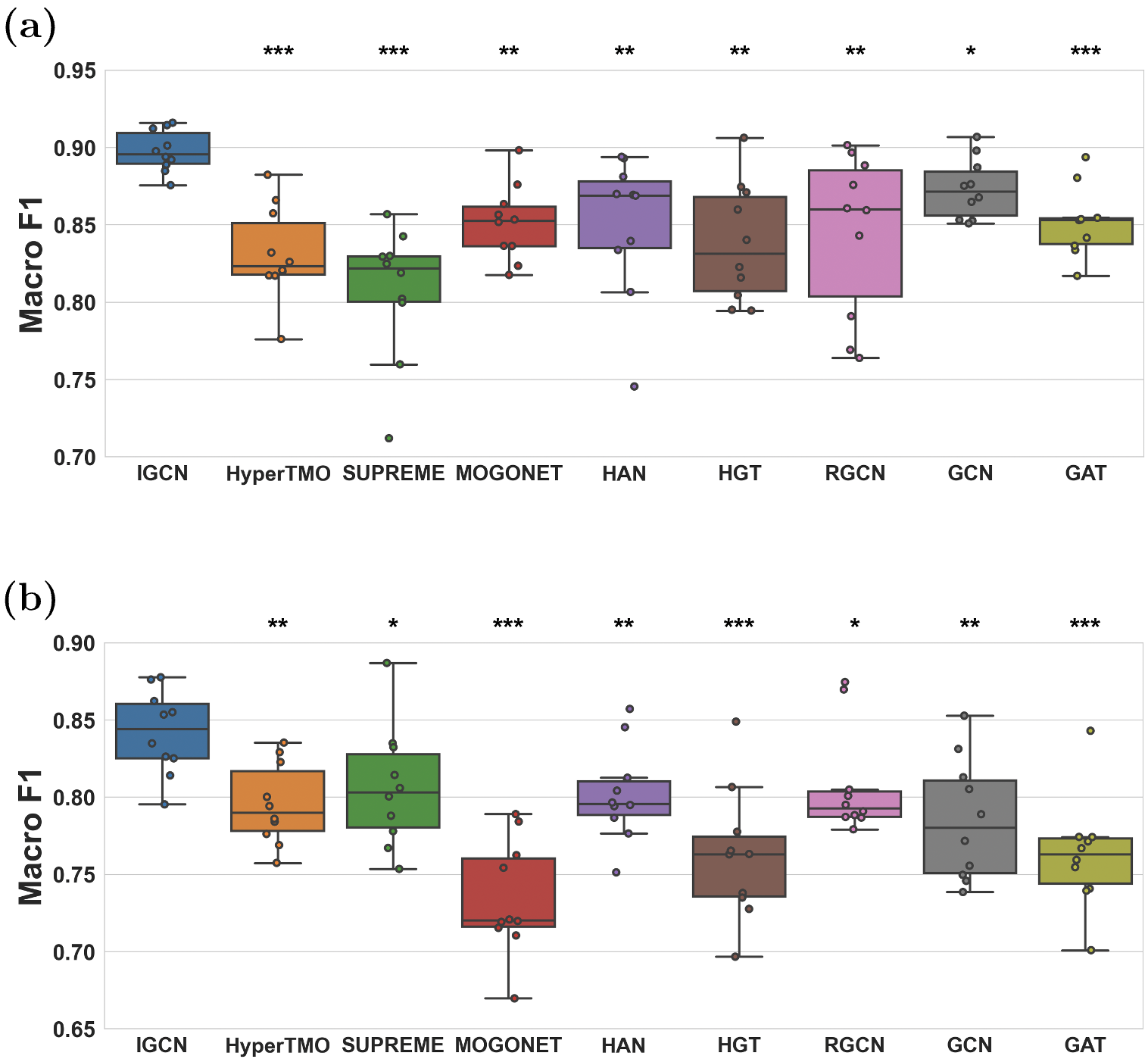}
    \caption{The boxplots show the distribution of macro F1 scores of ten different runs on \textbf{(a)} TCGA-GBM and \textbf{(b)} ADNI datasets for all methods. The means and standard deviations of these runs are shown in \textbf{Table~\ref{tab:ex_res}}. Wilcoxon rank-sum test p-values were computed between IGCN and other methods to compare the distribution of
box plots, representing p-value $<$ 0.001 by ***, else if $<$ 0.01 by **, and else if $<$ 0.05 by *.}
    \label{fig:box_plots}
\end{figure}

In \textbf{Fig.~\ref{fig:box_plots}(a)} and~\ref{fig:box_plots}\textbf{(b)}, the boxplots show the distribution of macro F1 scores of ten runs on GBM and ADNI datasets. \textbf{Table~\ref{tab:ex_res}} presents the means and standard deviations of these runs. We also calculated Wilcoxon rank-sum test p-values to compare the distribution of the box plots between IGCN and other methods. For TCGA-GBM, ADNI, and  TCGA-BRCA datasets, as shown in \textbf{Fig.~\ref{fig:box_plots}} and \textbf{Fig. S1(b)}, IGCN outperforms all other methods significantly (p-value $<$ 0.05).
For ROSMAP dataset, as shown in \textbf{Fig. S1(b)}, we observed statistically significant difference with all the methods (p-value $<$ 0.05) except for HyperTWO (p-value $>$ 0.05).

\begin{table*}[!]
    \centering
    \resizebox{1\textwidth}{!}{%
    \begin{tabular}{c|c|c|c|c|c}
       \textbf{Dataset}&\textbf{Method}  & \textbf{Accuracy} & \textbf{ Weighted F1} & \textbf{Macro F1} & \textbf{MCC}\\\hline
       \multirow{12}*{TCGA-BRCA}&MLP& $0.761 \pm 0.012$ &$ 0.752 \pm 0.015$ & $0.704 \pm 0.022$&$0.648 \pm 0.020$ \\
       &SVM& $0.774 \pm 0.022$ &$ 0.771 \pm 0.022$ & $0.736 \pm 0.027$&$0.668 \pm 0.033$ \\
       &RF& $0.714 \pm 0.020$ &$ 0.690 \pm 0.023$ & $0.594 \pm 0.036$&$0.565 \pm 0.032$ \\
       &GCN& $0.787 \pm 0.012$ &$ 0.782 \pm 0.016$ & $0.743 \pm 0.024$&$0.685 \pm 0.020$ \\ 
       &GAT& $0.789 \pm 0.015$ &$ 0.785 \pm 0.017$ & $0.747 \pm 0.022$&$0.688 \pm 0.024$ \\
       &HAN& $0.781 \pm 0.025$ &$ 0.772 \pm 0.036$ & $0.714 \pm 0.069$& $0.675 \pm 0.039$ \\ 
       &HGT& $0.795 \pm 0.028$ &$ 0.789 \pm 0.030$ & $0.739 \pm 0.044$&$0.697 \pm 0.042$ \\
       &RGCN& {$0.825 \pm 0.017$} &{$0.824 \pm 0.020$} & {$0.791 \pm 0.032$}&{$0.744\pm 0.027$} \\
       &MOGONET& $0.813 \pm 0.013$ &$ 0.813 \pm 0.014$ & $0.765 \pm 0.027$& $0.727 \pm 0.019$ \\
       &SUPREME& $0.821 \pm 0.020$ &$ 0.822 \pm 0.022$ & $0.783 \pm 0.032$&$0.742 \pm 0.031$ \\
       &HyperTMO& $\underline{0.838 \pm 0.015}$&$\underline{0.841 \pm 0.016}$ &$\underline{0.813 \pm 0.025}$&{$\underline{0.768 \pm 0.022}$} \\
       &IGCN& $\mathbf{0.874 \pm 0.011}$ &$ \mathbf{0.878 \pm 0.010}$ & $\mathbf{0.852 \pm 0.017}$& $\mathbf{0.821 \pm 0.014}$ \\\hline

       \multirow{12}*{TCGA-GBM}&MLP& $0.793 \pm 0.048$ &$ 0.791 \pm 0.050$ & $0.783 \pm 0.050$&$0.725 \pm 0.063$ \\
       &SVM& $0.779 \pm 0.057$ &$ 0.777 \pm 0.056$ & $0.767 \pm 0.058$&$0.706 \pm 0.075$ \\
       &RF& $0.818 \pm 0.049$ &$ 0.811 \pm 0.055$ & $0.801 \pm 0.059$&$0.761 \pm 0.063$ \\
       &GCN& $\underline{0.880 \pm 0.017}$ &$\underline{0.879 \pm 0.017}$ & $\underline{0.873 \pm 0.018}$&$\underline{0.839 \pm 0.023}$ \\ 
       &GAT& $0.861 \pm 0.018$ &$ 0.859 \pm 0.019$ & $0.852 \pm 0.021$&$0.814 \pm 0.025$ \\
       &HAN& $0.858 \pm 0.038$ &$ 0.856 \pm 0.041$ & $0.850 \pm 0.044$& $0.809 \pm 0.051$ \\ 
       &HGT& $0.840 \pm 0.033$ &$ 0.840 \pm 0.034$ & $0.839 \pm 0.036$&$0.788 \pm 0.045$ \\
       &RGCN& {$0.851 \pm 0.043$} &{$0.849 \pm 0.045$} & {$0.845 \pm 0.049$}&{$0.801 \pm 0.057$} \\
       &MOGONET& $0.854 \pm 0.019$ &$ 0.854 \pm 0.020$ & $0.851 \pm 0.023$& $0.805 \pm 0.026$ \\
       &SUPREME& $0.818 \pm 0.030$ &$ 0.816 \pm 0.034$ & $0.808 \pm 0.041$&$0.756 \pm 0.041$ \\
       &HyperTMO& ${0.837 \pm 0.026}$&${0.836 \pm 0.026}$ &${0.832 \pm 0.029}$&{${0.781 \pm 0.034}$} \\
       &IGCN& $\mathbf{0.903 \pm 0.014}$ &$ \mathbf{0.902 \pm 0.014}$ & $\mathbf{0.898 \pm 0.013}$& $\mathbf{0.870 \pm 0.019}$ \\\hline
       
       \multirow{12}*{ROSMAP}&MLP& $0.657 \pm 0.056$ &$ 0.650 \pm 0.059$ & $0.650 \pm 0.059$&$0.335 \pm 0.111$ \\
       &SVM& $0.645 \pm 0.063$ &$ 0.623 \pm 0.079$ & $0.621 \pm 0.081$&$0.308 \pm 0.131$ \\
       &RF& $0.692 \pm 0.066$ &$ 0.691 \pm 0.065$ & $0.691 \pm 0.066$&$0.387 \pm 0.136$ \\
       &GCN&{$0.701 \pm 0.042$} &{$0.700 \pm 0.042$} & {$0.699 \pm 0.042$}&${0.405 \pm 0.085}$ \\ 
       &GAT& $0.670 \pm 0.032$ &$ 0.669 \pm 0.033$ & $0.669 \pm 0.033$&$0.347 \pm 0.064$ \\
       &HAN& $0.775 \pm 0.025$ &$ 0.775 \pm 0.025$ & $0.774 \pm 0.025$& $0.550 \pm 0.052$ \\ 
       &HGT& $0.758 \pm 0.022$ &$ 0.756 \pm 0.022$ & $0.756 \pm 0.022$&$0.527 \pm 0.041$ \\
       &RGCN& $0.744 \pm 0.024$ &$ 0.741 \pm 0.024$ & $0.740 \pm 0.024$&$0.503 \pm 0.049$ \\
       &MOGONET& ${0.782 \pm 0.019}$ &${0.781 \pm 0.019}$ & ${0.781 \pm 0.019}$& $0.571 \pm 0.037$ \\
       &SUPREME& $0.782 \pm 0.028$&$0.781 \pm 0.028$ &$0.781 \pm 0.028$&{${0.575 \pm 0.056}$} \\
       &HyperTMO& $\underline{0.796 \pm 0.035}$&$\underline{0.795 \pm 0.035}$ &$\underline{0.795 \pm 0.035}$&{$\underline{0.596 \pm 0.071}$} \\
       &IGCN& $\mathbf{0.824 \pm 0.034}$ &$ \mathbf{0.823 \pm 0.034}$ & $\mathbf{0.823 \pm 0.034}$& $\mathbf{0.659 \pm 0.069}$ \\\hline
       \multirow{12}*{ADNI}&MLP& $0.774 \pm 0.028$ &$ 0.770 \pm 0.029$ & $0.770 \pm 0.029$&$0.660 \pm 0.041$ \\
       &SVM& $0.766 \pm 0.045$ &$ 0.763 \pm 0.048$ & $0.762 \pm 0.048$&$0.651 \pm 0.065$ \\
       &RF& $0.791 \pm 0.047$ &$ 0.788 \pm 0.048$ & $0.788 \pm 0.049$&$0.689 \pm 0.071
       $ \\
       &GCN& ${0.783 \pm 0.037}$ &${0.782 \pm 0.038}$ & ${0.785 \pm 0.037}$&${0.677 \pm 0.054}$ \\ 
       &GAT& $0.760 \pm 0.036$ &$ 0.759 \pm 0.035$ & $0.763 \pm 0.034$&$0.642 \pm 0.055$ \\
       &HAN& ${0.799 \pm 0.032}$ &$ {0.800 \pm 0.031}$ & ${0.802 \pm 0.029}$& ${0.702 \pm 0.047}$ \\ 
       &HGT& $0.758 \pm 0.039$ &$ 0.757 \pm 0.041$ & $0.762 \pm 0.041$&$0.642 \pm 0.056$ \\
       &RGCN& $\underline{0.807 \pm 0.034}$ &$\underline{0.806 \pm 0.034}$ & $\underline{0.808 \pm 0.033}$&$\underline{0.713 \pm 0.050}$ \\
       &MOGONET& $0.733 \pm 0.033$ &$ 0.732 \pm 0.034$ & $0.735 \pm 0.035$& $0.601 \pm 0.049$ \\
       &SUPREME& ${0.803 \pm 0.036}$ &${0.803 \pm 0.038}$ & ${0.806 \pm 0.037}$&${0.709 \pm 0.052}$ \\
       &HyperTMO& ${0.794 \pm 0.027}$&${0.793 \pm 0.025}$ &${0.796 \pm 0.025}$&{${0.695 \pm 0.042}$} \\
       &IGCN& {$\mathbf{0.840 \pm 0.026}$} &{$\mathbf{0.840 \pm 0.026}$} &{ $\mathbf{0.842 \pm 0.026}$}& {$\mathbf{0.762 \pm 0.039}$} \\\hline
    \end{tabular}
    }
    \caption{Classification results on TCGA-BRCA, TCGA-GBM, ROSMAP, and ADNI datasets. The reported values represent the averages along with standard deviations, based on ten runs, for four performance measures, namely: accuracy, macro F1, weighted F1, and Matthew's correlation coefficient (MCC). The best values for each dataset are shown in bold. The underline is used to signify the second-best performance.}
    \label{tab:ex_res}
\end{table*}

\subsection{Attention-driven Interpretability in IGCN}
\begin{figure}[b!]
    \centering
    \includegraphics[width=0.9\textwidth]{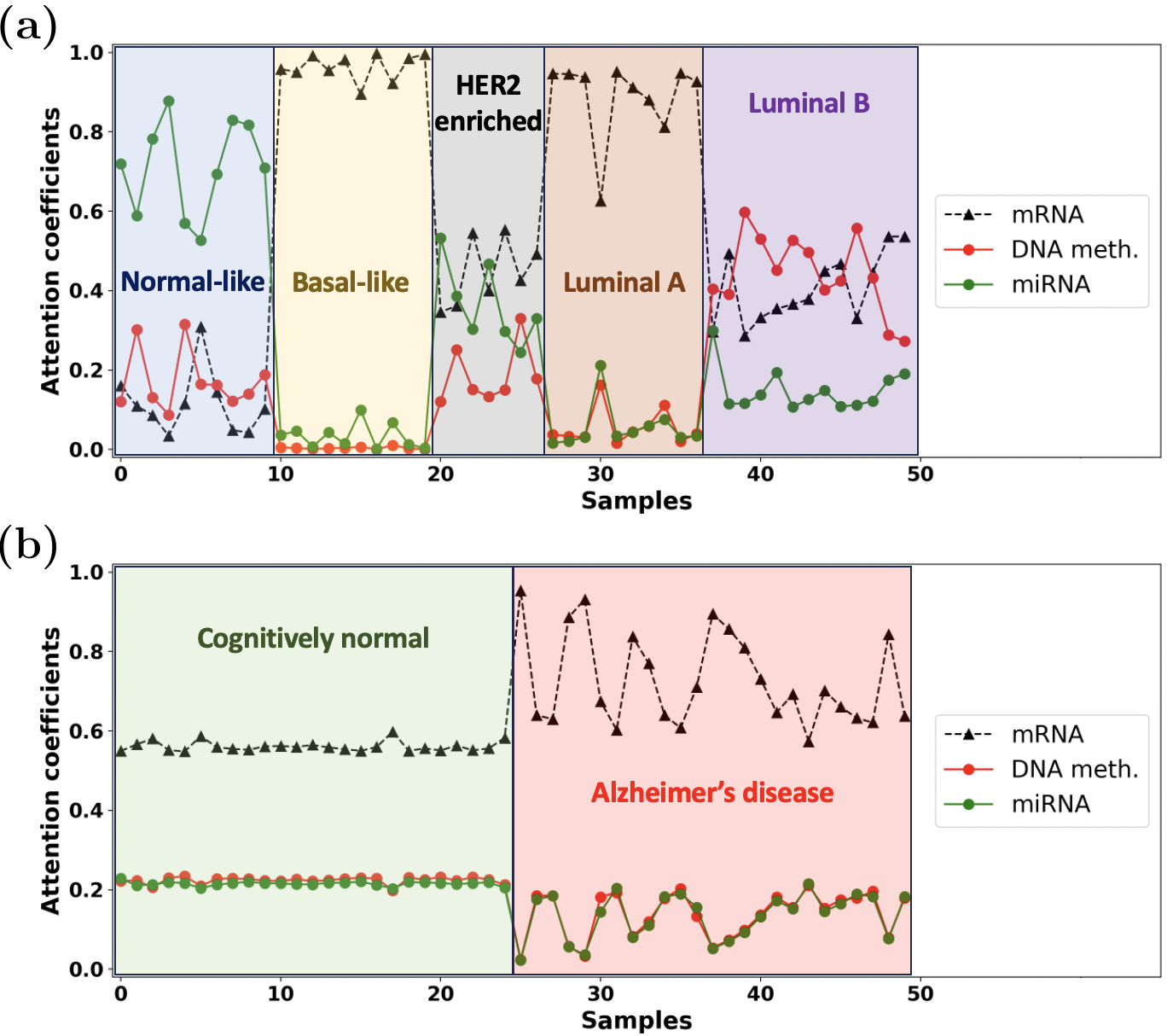}
    \caption{ Attention coefficients of 50 test samples of \textbf{(a)} TCGA-BRCA and \textbf{(b)} ROSMAP datasets. IGCN provides an attention mechanism, which computes a specific attention coefficient for each node embedding. 
    This speciality might allow us to investigate which feature is most informative for each sample in different node type prediction.}
    \label{fig:at}
\end{figure}

The personalized attention mechanism in IGCN is proposed to integrate multi-omics data embeddings into a weighted form. 
As this module assigns specific attention coefficients to each sample, we can observe unique characteristic pattern of each omics type at the label space. 
The significance of different embeddings derived from various types of omics data networks varies from sample to sample, depending on the cancer molecular type or specific diagnosis group \textbf{Fig.~\ref{fig:at}(a)} and~\ref{fig:at}\textbf{(b)} show the attention coefficients computed for 50 correctly predicted test samples in TCGA-BRCA and ROSMAP datasets, respectively.

For TCGA-BRCA dataset, mRNA expression data had the main contribution toward the prediction of Basal-like, HER2-enriched and Luminal A breast cancer subtypes, which is expected as PAM50 subtypes are based on mRNA expression data. Interestingly, however, miRNA expression data had the main contribution to  predicting Normal-like breast cancer subtype. It is also notable that the attention level of DNA methylation data is slightly higher compared to the attention level of mRNA expression data in the Luminal B samples. Concerning to ROSMAP dataset, mRNA expression plays a primary role in both CN and AD samples. However, the attention level given to mRNA expression data is slightly higher for AD samples compared to CN samples. The attention level of each omics type varies significantly across different samples. It is apparent that various types of omics data are integrated according to a patient-specific golden ratio. This feature makes IGCN interpretable in terms of understanding
the rationale behind the prediction at the
sample level. Moreover, it has the potential to pave the way for
a new research direction in analyzing different omics data types on different cancer subtype samples.

% \subsection{IGCN identifies significant biomarkers across various omics data types}
On the other hand, the feature ranking layer shown in \textbf{Fig.~\ref{fig:IGCN}} assigns attention coefficients to input features across all omics types. The feature ranking layer is unique such that it assigns attention values customized to each individual sample. This indicates that rather than employing a global attention mechanism that treats all samples uniformly, the module personalizes the attention scores for each individual sample. As illustrated in \textbf{Fig.~S2(a)}, the attention values for mRNA expressions within the TCGA-GBM dataset exhibit variability across samples. We enforce the normalization of attention values (given in Eq.~(\ref{eq:feature_att})), ensuring that their cumulative sum amounts to 1 for each sample. 
% For instance, in the TCGA-GBM dataset, each sample has 1,230 mRNA expressions as features. Consequently, a uniform attention value distributed across all expressions could be expressed as $\frac{1}{1,230}= 0.0008$. \SB{this part needs revision. reader doesn't know what you mean by the color bar. it's better to provide some interpretations without having them go to the supplemental file for this figure} While the color bar may suggest small attention values, it is important to note that these attention values might exhibit considerable or minor variations across the features \SB{you say considerable or minor variations. what is the message you want to give?} \CO{Let me explain once we meet}. 
Considering the attention scores across the features demonstrated in, it can be inferred that the genes with relatively higher attention scores hold greater significance compared to others. 
We ranked omics features based on their average attention across all samples and reported the top ten biomarkers in
\textbf{Table~\ref{tab:gene_ranks}}. Recent studies have provided evidence supporting the involvement of these biomarkers in the pathogenesis of GBM and AD~\cite{hares2017overexpression,hares2019kif5a,rocchio2019gene,liu2023early,hermkens2019profiling,li2021differentially,li2004analysis}.

In AD, beta-amyloid (A$\beta$) peptides can aggregate and accumulate, forming insoluble plaques in the brain. These plaques are a hallmark pathological feature of AD and are believed to contribute to the progressive neurodegeneration and cognitive decline seen in the disease. KIF5A gene is a protein-coding gene that belongs to the kinesin family. Some studies have suggested that KIF5A protein expression correlated inversely with the levels of soluble A$\beta$ in AD brains~\cite{hares2017overexpression,hares2019kif5a}. Research studies have suggested that CSRP1, HOPX, HMGN2, TF, and CDK2AP1 may play a role in the pathogenesis of AD through their involvement in gene regulation processes~\cite{rocchio2019gene,liu2023early,hermkens2019profiling,li2021differentially,li2004analysis}. Moreover, hsa-miR-27a-3p, hsa-miR-16-5p, hsa-miR-142-3p, hsa-miR-199a-5p, hsa-miR-107, and hsa-miR-1248 miRNAs have been identified as candidate biomarkers for AD~\cite{sala2013reduced,wang2022plasma,harati2022mir,swarbrick2019systematic,herrera2019systematic,kumar2013circulating,gattuso2022chronic,hewel2019common}. 

% The glyco-phosphoprotein osteopontin (OPN), encoded by SPP1 gene, has previously shown to be potentially associated with AD~\cite{comi2010osteopontin,chai2021plasma,de2023perivascular,hsu2024multi}. While SPARCL1 has been in various biological processes, some studies have suggested potential links between SPARCL1 and neuroinflammatory processes or synaptic plasticity, which are relevant to AD pathology~\cite{seddighi2018sparcl1,seddighi2018alpha2,pilozzi2020blood}. beta-amyloid (A$\beta$) clearance refers to the removal or degradation of A$\beta$ peptides from the brain. In AD, A$\beta$ peptides can aggregate and accumulate, forming insoluble plaques in the brain. These plaques are a hallmark pathological feature of AD and are believed to contribute to the progressive neurodegeneration and cognitive decline seen in the disease. Research studies have suggested that SERPINA3, CHI3L1, CST3, PMP2, and CRYAB may play a role in the pathogenesis of AD by affecting the clearance of A$\beta$~\cite{akbor2021polymorphic,shu2022detection,zeng2023astrocyte,blumenau2020investigating,olgiati2011genetics,bali2012role,montero2023proteomics,muraleva2021mek1}. Although there hasn't been extensive research on the relationship between PTPRZ1, HBB, and HBA2 genes and Alzheimer's disease (AD), some studies hint at potential connections between these genes and AD pathology~\cite{liu2024unraveling,fu2023blood,liu2022hydroxybenzyl}.

The genes SERPINA3, PTPRZ1, and CST3 have been extensively researched within the context of GBM. Serine protease inhibitor clade A, member 3 (SERPINA3) is a protein that influences GBM by modulating actions that promote tumor growth. SERPINA3 expression is higher in the peritumoral brain zone (PBZ) of GBM compared to the tumor core~\cite{giambra2023peritumoral}. The increased levels of SERPINA3 in the PBZ contribute to the aggressive behavior of GBM by facilitating the infiltration of tumor cells into surrounding brain tissue. This infiltration is a critical factor in the recurrence of the tumor, as the PBZ is often the site where tumor cells invade and spread into adjacent areas\cite{nimbalkar2021differential}. Elevated levels of PTPRZ1 are also associated with increased cell migration and invasion in GBM. Studies have demonstrated that downregulation of PTPRZ1 leads to reduced migration and invasion capabilities of glioma cells, suggesting that PTPRZ1 facilitates these processes, which are critical for tumor spread and recurrence~\cite{papadimitriou2023protein,xia2019expression,zeng2017tumour}. CST3 has been investigated as a potential prognostic marker in GBM. Elevated levels of CST3 may correlate with more aggressive tumor characteristics and poorer clinical outcomes, making it a candidate for further research in the context of GBM prognosis and treatment strategies~\cite{cheng2022cysteine,konduri2002modulation}. Furthermore, hsa-let-7b, hsa-let-7a, hsa-let-7c, and hsa-let-7f are members of the let-7 family of miRNAs, which have been extensively studied in various cancers, including GBM~\cite{li2016retracted,lee2011let}. These miRNAs have been implicated in regulating the expression of genes involved in GBM progression and may have potential as a therapeutic target~\cite{xi2019joint}.
hsa-miR-125b miRNA is part of a larger family of microRNAs that includes that includes miR-125b-1 and miR-125b-2~\cite{wang2020emerging}. Its upregulation can contribute to tumor growth and invasion by promoting pathways that enhance cell survival and proliferation~\cite{wang2020emerging}. hsa-miR-21 is known to be upregulated in GBM and has been implicated in promoting tumor growth, invasion, and resistance to therapy~\cite{akers2013mir}. hsa-miR-9 has been shown to regulate the mobility behavior of GBM cells and may have implications in GBM progression~\cite{ben2016hsa}.

% \begin{figure}[!]
%     \centering
%     \includegraphics[width=1\textwidth]{Fig/feature_ranking.png}
%     \caption{\textbf{(a)} The attention values for mRNA expressions and \textbf{(b)} the distribution of attention scores in TCGA-GBM dataset. The distribution of attention scores exhibits notable variations across the features.}
%     \label{fig:gene_ranks}
% \end{figure}

\begin{table*}[ht]
    \centering
     \resizebox{\textwidth}{!}{
    \begin{tabular}{c|c|c|c}
      \textbf{Dataset}   & \textbf{mRNA} & \textbf{miRNA} & \textbf{DNA methylation} \\\hline
        \multirow{10}*{TCGA-GBM} & \multicolumn{1}{p{3.7cm}}{\centering SSP1 \\ SPARCL1 \\SERPINA3\\ PTPRZ1 \\ CHI3L1 \\ CST3 \\ PMP2 \\ CRYAB \\ HBB \\ HBA2} & \multicolumn{1}{|p{3.5cm}|}{\centering hsa-let-7b \\ hsa-let-7a \\hsa-miR-125b\\ hsa-miR-21\\ hsa-miR-9 \\hsa-let-7c \\ hsa-let-7f \\hsa-miR-29a\\ hsa-miR-26a\\ hsa-miR-1290}& 
        \multicolumn{1}{p{3.5cm}}{\centering \multirow{10}{*}{N/A}} \\\hline
        \multirow{10}*{ROSMAP} & \multicolumn{1}{p{3.7cm}}{\centering QDPR \\ PPDPF\\PLEKHB1\\ KIF5A\\ CSRP1 \\ HOPX \\ HMGN2 \\PLEKHM2\\ TF \\ CDK2AP1 } & \multicolumn{1}{|p{3.5cm}|}{\centering hsa-miR-27a-3p \\ hsa-miR-340 \\hsa-miR-374b\\ hsa-miR-16-5p\\ hsa-miR-107 \\ hsa-miR-142-3p \\hsa-miR-199a-5p\\ hsa-miR-574-3p\\ hsa-miR-770-5p\\hsa-miR-1248}& 
        \multicolumn{1}{p{3.5cm}}{\centering cg02595219 \\ cg14837165 \\cg15775914\\ cg01182697\\ cg05382123\\cg07546360\\ cg12120741 \\cg20870559\\ cg23571857\\ cg24322623} \\
        % \multicolumn{1}{p{3.5cm}}{\centering KCNE3\\ D4ST1\\ CHML\\ TMEM59\\ CSMD2\\ FLJ27365\\ EDNRB\\ OAS2\\ BIRC4BP\\ MYOD1} \\
        \hline    
    \end{tabular}}
    \caption{The highest-ranked 10 omics biomarkers identified through IGCN in TCGA-GBM and ROSMAP datasets.}
    \label{tab:gene_ranks}
\end{table*}

\subsection{Ablation study}
We carried out an ablation study to investigate how the attention mechanism and Normalized Temperature-scaled Cross Entropy Loss ($\mathcal{L}_{NT-Xent}$) funtions affect the modeling ability of IGCN. Particularly, we developed three variants of IGCN: 1) we disabled the attention mechanism, and computed node embeddings as an average of node embeddings from each network; 2) we disabled the $\mathcal{L}_{NT-Xent}$ loss functions for all omics networks (shown in \textbf{Fig.~\ref{fig:IGCN}}); and 3) we used the proposed IGCN architecture to observe how disabling of different components affect the model performance. We conducted the experiments on TCGA-BRCA and ROSMAP datasets and reported the classification performance with macro F1 score measure in \textbf{Table~\ref{tab:ablation}}. The results show that both the attention mechanism module and the $\mathcal{L}_{NT-Xent}$ loss functions are vital in node classification tasks, as the full IGCN setup achieved the best performance. Therefore, directly combining different types of omics data embeddings without accounting for the sample-level importance of various networks, provided by the attention module, may lead to incorrect predictions and degrade performance. Additionally, some edges in the graph may not accurately represent the true interactions or relationships between nodes. These inaccuracies can adversely affect classification tasks. The results demonstrate that the $\mathcal{L}_{NT-Xent}$ loss can alleviate these inaccuracies and boost the classification performance of IGCN. 

\begin{table}[h!]
    \centering
    \resizebox{\textwidth}{!}{%
    \begin{tabular}{c|c|c|c} \multicolumn{2}{c|}{\textbf{Components}}&\multicolumn{2}{|c}{\textbf{Macro F1}}\\\hline
    Attn. Mech.&$\mathcal{L}_{NT-Xent}$&TCGA-BRCA&ROSMAP\\\hline
    \xmark & \cmark & $0.829\pm 0.023$&$0.750\pm 0.021$\\
    \cmark & \xmark & $0.831\pm 0.018$&$0.749\pm 0.035$\\
    \cmark & \cmark & $\mathbf{0.852\pm 0.017}$&$\mathbf{0.811\pm 0.032}$\\\hline
    \end{tabular}}
    \caption{The average macro F1 scores of different variants of IGCN on TCGA-BRCA and ROSMAP datasets over ten runs. Attn. Mech.: Attention mechanism, $\mathcal{L}_{NT-Xent}$:  Normalized Temperature-scaled Cross Entropy Loss.}
    \label{tab:ablation}
\end{table}

\section{Discussion}
\label{sec:sec3}
Due to the rapid advancements in omics technologies, along with major studies such as TCGA, ROSMAP, and ADNI, multi-omics datasets have become prevalent. Therefore, creating computational tools for the integrative analysis of various omics data types has been critically important in cancer molecular biology and precision medicine research. Toward 
this goal, to advance personalized medicine and uncover previously unknown characteristics through integrative analysis of multi-omics data, we present IGCN as a framework to integrate multi-omics datasets. To demonstrate the superiority of IGCN, we not only compare its performance with other multi-omics integrative tools but also compare it with recently introduced multi-modal graph representation learning methods, which have not been applied for multi-omics integration.

IGCN utilizes multi-GCN modules to extract node embeddings from each omics data. The personalized attention mechanism in IGCN is designed to integrate multi-omics data embeddings into a weighted form. This module assigns omic type attention values specific
for each sample, allowing us to observe the unique characteristic patterns of each omics type within the label space. Our findings revealed that the importance of different embeddings from various omics data networks changes for each sample~(\textbf{Fig. \ref{fig:at}}). In line with previous research, where mRNA expression data has been consistently recognized as a key factor in identifying breast cancer subtypes, our analysis of the TCGA-BRCA dataset confirms that mRNA expression significantly contributed to the prediction of Basal-like, HER2-enriched, and Luminal A subtypes. This is expected, given that PAM50 subtypes are traditionally defined based on mRNA expression profiles. Notably, however, our findings also reveal that miRNA expression data was the most influential in predicting the Normal-like breast cancer subtype, which adds a new dimension to the understanding of breast cancer classification and highlights the potential of miRNA as a complementary datatype in subtype prediction.  

% This observation aligns with emerging studies that suggest miRNAs play a critical role in the regulation of gene expression and could be pivotal in distinguishing specific cancer subtypes, particularly those with less defined mRNA expression patterns \SB{this sentence does not really say anything substantial. we should remove this}.

% \textcolor{green}{In alignment with previous research highlighting the significance of mRNA expression in neurodegenerative diseases, our analysis of  ROSMAP dataset confirms that mRNA expression plays a central role in both CN and AD samples. This is consistent with studies that have demonstrated the critical involvement of mRNA expression patterns in the molecular characterization of AD. Interestingly, we found that mRNA expression data receives slightly more attention in AD samples than in CN samples. This suggests that mRNA alterations play a more significant role in the progression of Alzheimer's disease, supporting recent studies that highlight the impact of mRNA regulation in AD pathogenesis.}\SB{these sentences in green are also not saying somethinig informative/insightful. mRNA expression represents the gene activity, which is important for everything. The following part is more important where you talk about specific genes or micRNAs. my point is that you can't say much when a particular omics type is important for a group of samples, especially if it is gene expression.}

Moreover, IGCN's ability to assign attention coefficients to features within individual samples across various omics data types represents a significant advancement. This capability not only aids in identifying key biomarkers in the pathogenesis of diseases such as BRCA, GBM, and AD but also suggests specific genes as candidates for predicting and differentiating cancer subtypes and other biomedical outcomes at the molecular level (\textbf{Table \ref{tab:gene_ranks}} and \textbf{Table S2}).  By effectively highlighting biomarkers across different omics layers, IGCN enhances our understanding of the underlying molecular mechanisms driving disease heterogeneity. This insight supports more accurate subtype classification, ultimately leading to more personalized and targeted therapeutic strategies.

% KIF5A gene, part of the kinesin family, has gained attention in Alzheimer's Disease research due to its potential involvement in modulating the dynamics of beta-A$\beta$ peptides, which are central to AD pathology~\cite{hares2017overexpression,hares2019kif5a}. IGCN has identified KIF5A as one of the top ten biomarkers in mRNA expression data\textbf{Table~\ref{tab:gene_ranks}}. Additionally, among our top ten mRNA biomarkers, we highlight hsa-let-7b, hsa-let-7a, hsa-let-7c, and hsa-let-7f miRNAs, which are members of the let-7 family. These miRNAs are known to play a role in regulating the expression of genes that are involved in the progression of GBM~\cite{li2016retracted,lee2011let}. 

% IGCN demonstrates the ability to assign attention coefficients to features within individual samples across various omics data types, enabling the identification of significant biomarkers from these diverse datasets. This feature enhances its utility in uncovering critical biomarkers involved in the pathogenesis of diseases such as BRCA, GBM, and AD, as supported by recent studies. 

One challenge encountered in our study was the limited availability of data across all modalities for some patients, which resulted in a smaller sample size. However, as more advanced datasets with additional omics layers become available, future studies could conduct a more comprehensive analysis. Such enriched data would allow for a deeper exploration of the complex interactions between different biological layers, ultimately enhancing our ability to uncover meaningful insights and improve predictive accuracy. Another important aspect to consider in our study is the use of datasets derived from bulk tissue samples rather than single-cell data. While bulk tissue analysis offers valuable insights into overall gene expression patterns, it can obscure the cellular heterogeneity within the tissue. Single-cell omics datasets would provide a more detailed view of the diversity among individual cells, enabling a deeper understanding of the specific roles of different cell populations in pathophysiology.

% \SB{you can say this earlier, but reduce the tone a bit. it sounds like commercial advertisement :) you should end your discussion about some future directions and possibly some limitations of the current study. for instance you could say the datasets were bulk tissue samples, not single cell. sample size was not great and some potential confounders were not available. with more advanced data with more omics layers, we could do more comprehensive analysis, etc.}Comparing to existing multi-omics integration tools, IGCN not only delivers superior performance but also provides enhanced interpretability, setting it apart in the field. Due to its innovative specialities, IGCN stands out as an innovative graph representation learning-based multi-omics integration approach for cancer subtype
% and biomedical classification applications, offering both exceptional performance and effective interpretability.

% Due to its innovative specialities, IGCN stands out as an innovative graph representation learning-based multi-omics integration approach for cancer subtype
% and biomedical classification applications, offering both exceptional performance and effective interpretability. To the best of our knowledge,  IGCN  represents a breakthrough as the first supervised integrative approach offering patient-level insights and biomarkers in multi-omics integration.

\section{Methods}
\label{sec:sec4}
\subsection{Customizing GCN for omics-focused learning}
IGCN employs GCN modules on graph networks to obtain the node embeddings.  Each GCN module in IGCN can be defined as:
\begin{equation}
\label{eq:node_emb}
    H_i = \sigma(D_i^{-1/2}A_i D_i^{-1/2}X_iW_i),
\end{equation}
for $i=1,2,...,p$, where $p$ is the total number of data modalities (omics types) and $X_i \in \mathbb{R}^{m \times d}$ is the feature matrix ($m$ is the number of nodes and $d$ is the feature size). $D_i$ and $W_i$ are the node degree and the learnable weight matrices, respectively. $\sigma$ is the activation function. We used a single layer GCN to obtain the node embeddings ($H_i$) for each network layer, however, a multi-layer GCN can be considered as outlined in~\citep{kipf2016semi}. 

In our work, the original adjacency matrix ${A}_i \in \mathbb{R}^{m \times m}$ was constructed by calculating
cosine similarity of each node pair and filtering out edges with cosine similarity $<\epsilon$. The adjacency matrix can be defined as:
% \begin{equation}
% \label{eq:adj_mtx}
%     \Hat{A}_i=
%     \begin{cases}
%       Ind(s(\mathbf{x}_i^q,\mathbf{x}_i^w)), & \text{if}\;\ q\neq w \;\text{and}\; s(\mathbf{x}_q,\mathbf{x}_w) \geq \epsilon \\
%       0, & \text{otherwise}
%     \end{cases}
% \end{equation}
\begin{equation}
\label{eq:adj_mtx}
    {a}_i^{(q,w)}=
      Ind(s(\mathbf{x}_i^q,\mathbf{x}_i^w)),
\end{equation}
where $\mathbf{x}_i^q$ and $\mathbf{x}_i^w$ are the node features of the node $q$ and $w$ (the $q^{th}$ and $w^{th}$ row vectors of $X_i$), respectively. ${a}_i^{(q,w)}$ is an element of the matrix ${A}_i$ corresponding to the $q^{th}$ row and $w^{th}$ column. 
$s(\mathbf{x}_i^q,\mathbf{x}_i^w)=\frac{\langle \mathbf{x}_i^q, \mathbf{x}_i^w\rangle}{||\mathbf{x}_i^q||_2||\mathbf{x}_i^w||_2}$ is the cosine similarity. $Ind(.)$ is an indicator function that maps the input to 1 if the input is greater than or equal to $\epsilon$, and to 0 otherwise. As shown in \textbf{Algorithm~\ref{<alg-label>}}, the threshold $\epsilon$ can be determined based on a given parameter $k$ as:
\begin{equation}
\label{eq:treshold}
    k = \frac{1}{m}\sum_{q,w} Ind(s(\mathbf{x}_i^q,\mathbf{x}_i^w)).
\end{equation}

\begin{algorithm}
\caption{Determine threshold $\epsilon$ based on pre-specified parameter $k$}\label{<alg-label>}
\begin{algorithmic}[1]
\State {\textbf{Input:} pre-specified parameter $k$ and initialize $\epsilon=1$ }
\State {\textbf{Output:} $\epsilon$}
\While{\textbf{true}}
    \State $counter = \frac{1}{m}\sum_{q,w} Ind(s(\mathbf{x}_i^q,\mathbf{x}_i^w))$
    \If{$counter>=k$}
    \State \textbf{break}
    \Else
    \State{$\epsilon \leftarrow \epsilon - 10^{-6}$}
    \EndIf
\EndWhile
\end{algorithmic}
\end{algorithm}
\noindent Choosing a proper $k$ value depends on the topological structure of the data. The results shown in \textbf{Fig. S3} indicate that IGCN was robust to the change of $k$ value and outperformed other integrative tools under different $k$ values. 

When building a graph based on features and employing cosine similarity, it is frequent to come across data noise arising from diverse origins, such as measurement inaccuracies, missing values, or inherent fluctuations within the dataset. These inaccuracies can have negative impacts on node classification tasks. To alleviate these inaccuracies, the Normalized Temperature-scaled Cross Entropy loss (NT-Xent loss)~\cite{sohn2016improved} was utilized for each GCN module.
\begin{align}
    pos_q = \sum_{j \in +}exp(s(\mathbf{x}_i^q,\mathbf{x}_i^j))/\tau,\label{eq:pos_q}\\
    neg_q = \sum_{\ell \in -}exp(s(\mathbf{x}_i^q,\mathbf{x}_i^{\ell}))/\tau.\label{eq:neg_q}
\end{align}

In Eq.~(\ref{eq:pos_q}), we calculate the cosine similarity between the anchor sample $q$ and its positive pairs, scaled by the temperature parameter $\tau$. Similarly, in Eq.~(\ref{eq:neg_q}), we calculate the cosine similarity between the anchor sample $q$ and its negative pairs, scaled by the temperature parameter. It is noteworthy that the anchor sample and its corresponding positive pairs belong to the same class, while the negative pairs were selected from classes different from that of the anchor sample. Thus, to learn representations that bring similar data points closer in the embedding space while pushing dissimilar data points farther apart, the NT-Xent loss for the $i^{th}$ GCN module can be defined as follows:
\begin{equation}
\label{eq:NT-xent}
       \mathcal{L}_{i_{NT-Xent}} = \frac{1}{\upsilon}\sum_{q=1}^{\upsilon} log\Bigg( \frac{pos_q}{pos_q + neg_q}\Bigg).
\end{equation}
where $\upsilon$ is the total number of samples in the training set.
\subsection{Computing attention coefficients and predictions}
\label{subsec:2}

After computing node embeddings using Eq. (\ref{eq:node_emb}), IGCN provides an attention mechanism to fuse multiple node embeddings into a weighted form by assigning attention coefficients to node embeddings. Inspired from ~\citep{velivckovic2017graph,wang2019heterogeneous}, attention coefficients can be determined as:
\begin{equation}
    \label{eq:at_coef}
    c_i^n = \frac{exp(LeakyReLU(\mathbf{h}_i^nW_a + b))}{\sum_{j =1} ^{p} exp(LeakyReLU(\mathbf{h}_j^nW_a + b))} ,
\end{equation}
where $\mathbf{h}_i^n \in \mathbb{R}^{d}$ is the $n^{th}$ node embedding of the $i^{th}$ similarity network and $p$ is the total number of data modalities. $W_a \in \mathbb{R}^{d \times 1}$ and $b \in \mathbb{R}^{1}$ are learnable weight and bias parameters, respectively.  Attention coefficient $c_i^n \in \mathbb{R}^{1}$ represents the importance of the $n^{th}$ node embedding of the $i^{th}$ network. Attention coefficients can be computed for all nodes of the similarity network and represented as a vector. Therefore, we can fuse the multiple node embeddings using element-wise multiplication, as follows:
\begin{equation}
    \label{eq:w_form}
    Z = \sum_{i=1}^p\Vec{\mathbf{c}_i}\otimes H_i,
\end{equation}
where ``$\otimes$'' denotes element-wise multiplication. $H_i$ is the node embedding matrix corresponding to $i^{th}$ similarity network and 
$\Vec{\mathbf{c}_i}$ is the attention coefficient vector for the nodes in the $i^{th}$ similarity network. It conveys that, although all embeddings are derived from a particular network, individual node embeddings may have distinct coefficient values. For example, consider the vector $\Vec{\mathbf{c}_1}$ as a column vector of size $m$:
\begin{equation*}
\Vec{\mathbf{c}_1} =
    \begin{bmatrix}
        c_{1}^{1}\\c_{1}^{2}\\ \vdots\\c_{1}^{m}
    \end{bmatrix},
\end{equation*}
and let $H_1$ be a matrix of size $m \times d$:
\begin{equation*}
H_1 =
    \begin{bmatrix}
        h_1^{(1,1)}&h_1^{(1,2)}&\hdots&h_1^{(1,d)}\\ h_1^{(2,1)}&h_1^{(2,2)}&\hdots&h_1^{(2,d)}\\
        \vdots&\vdots&\ddots&\vdots
        \\
        h_1^{(m,1)}&h_1^{(m,2)}&\hdots&h_1^{(m,d)}
    \end{bmatrix}.
\end{equation*}
We also note that $n^{th}$ row of $H_1$ can also be represented as a row vector $\mathbf{h}_1^n$:
\begin{equation*}
\mathbf{h}_1^n =
    \begin{bmatrix}
        h_1^{(n,1)}&h_1^{(n,2)}&\hdots&h_1^{(n,d)}
    \end{bmatrix}.
\end{equation*}
The element-wise multiplication of each row of $\Vec{\mathbf{c}_1}$ by the corresponding row of $H_1$ can be represented as follows:
\begin{equation*}
\Vec{\mathbf{c}_1}\otimes 
H_1 =
    \begin{bmatrix}
        c_{1}^{1}\cdot h_1^{(1,1)}&c_{1}^{1}\cdot h_1^{(1,2)}&\hdots&c_{1}^{1}\cdot h_1^{(1,d)}\\ c_{1}^{2}\cdot h_1^{(2,1)}&c_{1}^{2}\cdot h_1^{(2,2)}&\hdots&c_{1}^{2}\cdot h_1^{(2,d)}\\
        \vdots&\vdots&\ddots&\vdots
        \\
        c_{1}^{m}\cdot h_1^{(m,1)}&c_{1}^{m}\cdot h_1^{(m,2)}&\hdots&c_{1}^{m}\cdot h_1^{(m,d)}
    \end{bmatrix}.
\end{equation*}

The weighted form of embeddings computed using Eq. (\ref{eq:w_form}) is utilized on a neural network to obtain the node label predictions. Thus, it can be written as:
\begin{equation}
    \label{eq:pred}
    \Hat{Y} = \sigma(Z\Bar{W}_1)\Bar{W}_2,
\end{equation}
where $\Bar{W}_1$ and $\Bar{W}_2$ are  learnable weight matrices. 

Besides the NT-Xent loss given in Eq.~(\ref{eq:NT-xent}), we also used the cross entropy loss as follows:
\begin{equation}
    \mathcal{L}_{CE} = \sum_{j=1}^v-log\Bigg(\frac{e^{\langle \hat{\mathbf{y}}^j, \mathbf{y}^j\rangle}}{\sum_k e^{\hat{{y}}^{(j,k)}}}\Bigg),
\end{equation}
where $\hat{\mathbf{y}}^j \in \mathbb{R}^{d}$ is the $j^{th}$ row in $\Hat{Y}$, which is the predicted label distribution of the $j^{th}$ training sample. $\hat{{y}}^{(j,k)}$ is the $k^{th}$ element in $\hat{\mathbf{y}}^j$. $\mathbf{y}^j$ is the one-hot encoded vector of the ground truth label of the $j^{th}$ training sample. $\langle \hat{\mathbf{y}}^j, \mathbf{y}^j\rangle$
represents the inner product of the vector $\hat{\mathbf{y}}^j$ and the vector $\mathbf{y}^j$. 
To determine all learnable weights and biases, the total lost function can be written as:
\begin{equation}
    \mathcal{L}_{IGCN} = \mathcal{L}_{CE} + \sum_{i=1}^p \mathcal{L}_{i_{NT-Xent}},
\end{equation}
where $p$ is the total number of data modalities.

Adam optimization~\citep{kingma2014adam} was used as the state-of-the-art for stochastic gradient descent algorithm and 0.5 dropout was added for each GCN layer. Early stopping
was used with the patience of 30 forced to have at least 200 epochs, which were determined empirically.

\subsection{Uncovering significant biomarkers during the prediction process}
Another component of IGCN is the feature ranking module which identifies noteworthy biomarkers across various omics datasets. Similar to the attention mechanism module given in Sec.~\ref{subsec:2}, the feature ranking module also offers personalized insights regarding feature importance. The attention values for each feature can be computed as follows:
\begin{equation}
    \label{eq:feature_att}
    r_i^{(j,k)} = \frac{exp\Bigg(LeakyReLU\Big(\big(LeakyReLU(x_i^{(j,k)}\Tilde{W}_i)\big)\Hat{W}_i + \Hat{b}_i\Big)\Bigg)}{\sum_{k =1} ^{d_i} exp\Bigg(LeakyReLU\Big(\big(LeakyReLU(x_i^{(j,k)}\Tilde{W}_i )\big)\Hat{W}_i + \Hat{b}_i\Big)\Bigg)},
\end{equation}
where $r_i^{(j,k)} \in \mathbb{R}^1$ represents the attention value of the $k^{th}$ feature in the $i^{th}$ omics type corresponding to the $j^{th}$ sample.  $x_i^{(j,k)}  \in \mathbb{R}^1 $ is the $k^{th}$ row feature in the $i^{th}$ omics type corresponding to the $j^{th}$ sample. $\Tilde{W}_i \in \mathbb{R}^{1 \times \omega}$ and $\Hat{W}_i \in \mathbb{R}^{\omega \times 1}$ are the learnable weight matrices. $\Hat{b}_i \in \mathbb{R}^1$ is the bias parameter. As the input $x_i^{(j,k)}$ is a scalar, $\Tilde{W}_i \in \mathbb{R}^{1 \times \omega}$ and $\Hat{W}_i \in \mathbb{R}^{\omega \times 1}$ were employed as an expansion and a compression units, respectively. $d_i$ is the feature size of the $i^{th}$ omics type and $\omega$ is the latent space dimension. Hence, we assign a feature rank for each feature used as an input. Moreover, since each sample was evaluated individually, the significance of each feature may differ for each patient.

\section{Acknowledgements}
This work was supported by the National Institute of General Medical Sciences of the National Institutes of Health under Award Number R35GM133657. Data collection and sharing for this project was funded by ADNI (National Institutes of Health Grant U01 AG024904) and DOD ADNI (Department of Defense award number W81XWH-12-2-0012).  
ADNI is funded by the National Institute on Aging, the National Institute of Biomedical Imaging and Bioengineering, and through generous contributions from the following: AbbVie, Alzheimer’s Association; Alzheimer’s Drug Discovery Foundation; Araclon Biotech; BioClinica, Inc.; Biogen; Bristol-Myers Squibb Company; CereSpir, Inc.; Cogstate; Eisai Inc.; Elan Pharmaceuticals, Inc.; Eli Lilly and Company; EuroImmun; F. Hoffmann-La Roche Ltd and its affiliated company Genentech, Inc.; Fujirebio; GE Healthcare; IXICO Ltd.; Janssen Alzheimer Immunotherapy Research \& Development, LLC.; Johnson \& Johnson Pharmaceutical Research \& Development LLC.; Lumosity; Lundbeck; Merck \& Co., Inc.; Meso Scale Diagnostics, LLC.; NeuroRx Research; Neurotrack Technologies; Novartis Pharmaceuticals Corporation; Pfizer Inc.; Piramal Imaging; Servier; Takeda Pharmaceutical Company; and Transition Therapeutics. The Canadian Institutes of Health Research is providing funds to support ADNI clinical sites in Canada. Private sector contributions are facilitated by the Foundation for the National Institutes of Health (www.fnih.org). The grantee organization is the Northern California Institute for Research and Education, and the study is coordinated by the Alzheimer’s Therapeutic Research Institute at the University of Southern California. ADNI data are disseminated by the Laboratory for Neuro Imaging at the University of Southern California.

\section{Code availability}
The source code and documentation is available at \url{https://github.com/bozdaglab/IGCN}.


\bibliography{sn-article}% common bib file


\begin{thebibliography}{59}
\ifx \bisbn   \undefined \def \bisbn  #1{ISBN #1}\fi
\ifx \binits  \undefined \def \binits#1{#1}\fi
\ifx \bauthor  \undefined \def \bauthor#1{#1}\fi
\ifx \batitle  \undefined \def \batitle#1{#1}\fi
\ifx \bjtitle  \undefined \def \bjtitle#1{#1}\fi
\ifx \bvolume  \undefined \def \bvolume#1{\textbf{#1}}\fi
\ifx \byear  \undefined \def \byear#1{#1}\fi
\ifx \bissue  \undefined \def \bissue#1{#1}\fi
\ifx \bfpage  \undefined \def \bfpage#1{#1}\fi
\ifx \blpage  \undefined \def \blpage #1{#1}\fi
\ifx \burl  \undefined \def \burl#1{\textsf{#1}}\fi
\ifx \doiurl  \undefined \def \doiurl#1{\url{https://doi.org/#1}}\fi
\ifx \betal  \undefined \def \betal{\textit{et al.}}\fi
\ifx \binstitute  \undefined \def \binstitute#1{#1}\fi
\ifx \binstitutionaled  \undefined \def \binstitutionaled#1{#1}\fi
\ifx \bctitle  \undefined \def \bctitle#1{#1}\fi
\ifx \beditor  \undefined \def \beditor#1{#1}\fi
\ifx \bpublisher  \undefined \def \bpublisher#1{#1}\fi
\ifx \bbtitle  \undefined \def \bbtitle#1{#1}\fi
\ifx \bedition  \undefined \def \bedition#1{#1}\fi
\ifx \bseriesno  \undefined \def \bseriesno#1{#1}\fi
\ifx \blocation  \undefined \def \blocation#1{#1}\fi
\ifx \bsertitle  \undefined \def \bsertitle#1{#1}\fi
\ifx \bsnm \undefined \def \bsnm#1{#1}\fi
\ifx \bsuffix \undefined \def \bsuffix#1{#1}\fi
\ifx \bparticle \undefined \def \bparticle#1{#1}\fi
\ifx \barticle \undefined \def \barticle#1{#1}\fi
\bibcommenthead
\ifx \bconfdate \undefined \def \bconfdate #1{#1}\fi
\ifx \botherref \undefined \def \botherref #1{#1}\fi
\ifx \url \undefined \def \url#1{\textsf{#1}}\fi
\ifx \bchapter \undefined \def \bchapter#1{#1}\fi
\ifx \bbook \undefined \def \bbook#1{#1}\fi
\ifx \bcomment \undefined \def \bcomment#1{#1}\fi
\ifx \oauthor \undefined \def \oauthor#1{#1}\fi
\ifx \citeauthoryear \undefined \def \citeauthoryear#1{#1}\fi
\ifx \endbibitem  \undefined \def \endbibitem {}\fi
\ifx \bconflocation  \undefined \def \bconflocation#1{#1}\fi
\ifx \arxivurl  \undefined \def \arxivurl#1{\textsf{#1}}\fi
\csname PreBibitemsHook\endcsname

%%% 1
\bibitem[\protect\citeauthoryear{Chaudhary et~al.}{2018}]{chaudhary2018deep}
\begin{barticle}
\bauthor{\bsnm{Chaudhary}, \binits{K.}},
\bauthor{\bsnm{Poirion}, \binits{O.B.}},
\bauthor{\bsnm{Lu}, \binits{L.}},
\bauthor{\bsnm{Garmire}, \binits{L.X.}}:
\batitle{Deep learning--based multi-omics integration robustly predicts survival in liver cancer}.
\bjtitle{Clinical Cancer Research}
\bvolume{24}(\bissue{6}),
\bfpage{1248}--\blpage{1259}
(\byear{2018})
\end{barticle}
\endbibitem

%%% 2
\bibitem[\protect\citeauthoryear{Poirion et~al.}{2018}]{poirion2018deep}
\begin{barticle}
\bauthor{\bsnm{Poirion}, \binits{O.B.}},
\bauthor{\bsnm{Chaudhary}, \binits{K.}},
\bauthor{\bsnm{Garmire}, \binits{L.X.}}:
\batitle{Deep learning data integration for better risk stratification models of bladder cancer}.
\bjtitle{AMIA Summits on Translational Science Proceedings}
\bvolume{2018},
\bfpage{197}
(\byear{2018})
\end{barticle}
\endbibitem

%%% 3
\bibitem[\protect\citeauthoryear{Sharifi-Noghabi et~al.}{2019}]{sharifi2019moli}
\begin{barticle}
\bauthor{\bsnm{Sharifi-Noghabi}, \binits{H.}},
\bauthor{\bsnm{Zolotareva}, \binits{O.}},
\bauthor{\bsnm{Collins}, \binits{C.C.}},
\bauthor{\bsnm{Ester}, \binits{M.}}:
\batitle{Moli: multi-omics late integration with deep neural networks for drug response prediction}.
\bjtitle{Bioinformatics}
\bvolume{35}(\bissue{14}),
\bfpage{501}--\blpage{509}
(\byear{2019})
\end{barticle}
\endbibitem

%%% 4
\bibitem[\protect\citeauthoryear{Huang et~al.}{2019}]{huang2019salmon}
\begin{barticle}
\bauthor{\bsnm{Huang}, \binits{Z.}},
\bauthor{\bsnm{Zhan}, \binits{X.}},
\bauthor{\bsnm{Xiang}, \binits{S.}},
\bauthor{\bsnm{Johnson}, \binits{T.S.}},
\bauthor{\bsnm{Helm}, \binits{B.}},
\bauthor{\bsnm{Yu}, \binits{C.Y.}},
\bauthor{\bsnm{Zhang}, \binits{J.}},
\bauthor{\bsnm{Salama}, \binits{P.}},
\bauthor{\bsnm{Rizkalla}, \binits{M.}},
\bauthor{\bsnm{Han}, \binits{Z.}}, \betal:
\batitle{Salmon: survival analysis learning with multi-omics neural networks on breast cancer}.
\bjtitle{Frontiers in genetics}
\bvolume{10},
\bfpage{166}
(\byear{2019})
\end{barticle}
\endbibitem

%%% 5
\bibitem[\protect\citeauthoryear{Choi and Chae}{2023}]{choi2023mobrca}
\begin{barticle}
\bauthor{\bsnm{Choi}, \binits{J.M.}},
\bauthor{\bsnm{Chae}, \binits{H.}}:
\batitle{mobrca-net: a breast cancer subtype classification framework based on multi-omics attention neural networks}.
\bjtitle{BMC bioinformatics}
\bvolume{24}(\bissue{1}),
\bfpage{169}
(\byear{2023})
\end{barticle}
\endbibitem

%%% 6
\bibitem[\protect\citeauthoryear{Khadirnaikar et~al.}{2023}]{khadirnaikar2023machine}
\begin{barticle}
\bauthor{\bsnm{Khadirnaikar}, \binits{S.}},
\bauthor{\bsnm{Shukla}, \binits{S.}},
\bauthor{\bsnm{Prasanna}, \binits{S.}}:
\batitle{Machine learning based combination of multi-omics data for subgroup identification in non-small cell lung cancer}.
\bjtitle{Scientific Reports}
\bvolume{13}(\bissue{1}),
\bfpage{4636}
(\byear{2023})
\end{barticle}
\endbibitem

%%% 7
\bibitem[\protect\citeauthoryear{Gong et~al.}{2023}]{gong2023multi}
\begin{barticle}
\bauthor{\bsnm{Gong}, \binits{P.}},
\bauthor{\bsnm{Cheng}, \binits{L.}},
\bauthor{\bsnm{Zhang}, \binits{Z.}},
\bauthor{\bsnm{Meng}, \binits{A.}},
\bauthor{\bsnm{Li}, \binits{E.}},
\bauthor{\bsnm{Chen}, \binits{J.}},
\bauthor{\bsnm{Zhang}, \binits{L.}}:
\batitle{Multi-omics integration method based on attention deep learning network for biomedical data classification}.
\bjtitle{Computer Methods and Programs in Biomedicine}
\bvolume{231},
\bfpage{107377}
(\byear{2023})
\end{barticle}
\endbibitem

%%% 8
\bibitem[\protect\citeauthoryear{Abbas et~al.}{2022}]{abbas2022alzheimer}
\begin{barticle}
\bauthor{\bsnm{Abbas}, \binits{Z.}},
\bauthor{\bsnm{Tayara}, \binits{H.}},
\bauthor{\bsnm{Chong}, \binits{K.T.}}:
\batitle{Alzheimer's disease prediction based on continuous feature representation using multi-omics data integration}.
\bjtitle{Chemometrics and Intelligent Laboratory Systems}
\bvolume{223},
\bfpage{104536}
(\byear{2022})
\end{barticle}
\endbibitem

%%% 9
\bibitem[\protect\citeauthoryear{Shigemizu et~al.}{2020}]{shigemizu2020prognosis}
\begin{barticle}
\bauthor{\bsnm{Shigemizu}, \binits{D.}},
\bauthor{\bsnm{Akiyama}, \binits{S.}},
\bauthor{\bsnm{Higaki}, \binits{S.}},
\bauthor{\bsnm{Sugimoto}, \binits{T.}},
\bauthor{\bsnm{Sakurai}, \binits{T.}},
\bauthor{\bsnm{Boroevich}, \binits{K.A.}},
\bauthor{\bsnm{Sharma}, \binits{A.}},
\bauthor{\bsnm{Tsunoda}, \binits{T.}},
\bauthor{\bsnm{Ochiya}, \binits{T.}},
\bauthor{\bsnm{Niida}, \binits{S.}}, \betal:
\batitle{Prognosis prediction model for conversion from mild cognitive impairment to alzheimer’s disease created by integrative analysis of multi-omics data}.
\bjtitle{Alzheimer's research \& therapy}
\bvolume{12},
\bfpage{1}--\blpage{12}
(\byear{2020})
\end{barticle}
\endbibitem

%%% 10
\bibitem[\protect\citeauthoryear{Li et~al.}{2021}]{li2021applied}
\begin{barticle}
\bauthor{\bsnm{Li}, \binits{Z.}},
\bauthor{\bsnm{Jiang}, \binits{X.}},
\bauthor{\bsnm{Wang}, \binits{Y.}},
\bauthor{\bsnm{Kim}, \binits{Y.}}:
\batitle{Applied machine learning in alzheimer's disease research: omics, imaging, and clinical data}.
\bjtitle{Emerging topics in life sciences}
\bvolume{5}(\bissue{6}),
\bfpage{765}--\blpage{777}
(\byear{2021})
\end{barticle}
\endbibitem

%%% 11
\bibitem[\protect\citeauthoryear{Wang et~al.}{2021}]{wang2021mogonet}
\begin{barticle}
\bauthor{\bsnm{Wang}, \binits{T.}},
\bauthor{\bsnm{Shao}, \binits{W.}},
\bauthor{\bsnm{Huang}, \binits{Z.}},
\bauthor{\bsnm{Tang}, \binits{H.}},
\bauthor{\bsnm{Zhang}, \binits{J.}},
\bauthor{\bsnm{Ding}, \binits{Z.}},
\bauthor{\bsnm{Huang}, \binits{K.}}:
\batitle{Mogonet integrates multi-omics data using graph convolutional networks allowing patient classification and biomarker identification}.
\bjtitle{Nature Communications}
\bvolume{12}(\bissue{1}),
\bfpage{3445}
(\byear{2021})
\end{barticle}
\endbibitem

%%% 12
\bibitem[\protect\citeauthoryear{Kesimoglu and Bozdag}{2023}]{kesimoglu2022supreme}
\begin{barticle}
\bauthor{\bsnm{Kesimoglu}, \binits{Z.N.}},
\bauthor{\bsnm{Bozdag}, \binits{S.}}:
\batitle{{SUPREME}: multiomics data integration using graph convolutional networks}.
\bjtitle{NAR Genomics and Bioinformatics}
\bvolume{5}(\bissue{2}),
\bfpage{063}
(\byear{2023})
\end{barticle}
\endbibitem

%%% 13
\bibitem[\protect\citeauthoryear{Yin et~al.}{2022}]{yin2022molecular}
\begin{barticle}
\bauthor{\bsnm{Yin}, \binits{C.}},
\bauthor{\bsnm{Cao}, \binits{Y.}},
\bauthor{\bsnm{Sun}, \binits{P.}},
\bauthor{\bsnm{Zhang}, \binits{H.}},
\bauthor{\bsnm{Li}, \binits{Z.}},
\bauthor{\bsnm{Xu}, \binits{Y.}},
\bauthor{\bsnm{Sun}, \binits{H.}}:
\batitle{Molecular subtyping of cancer based on robust graph neural network and multi-omics data integration}.
\bjtitle{Frontiers in Genetics}
\bvolume{13},
\bfpage{884028}
(\byear{2022})
\end{barticle}
\endbibitem

%%% 14
\bibitem[\protect\citeauthoryear{Xiao et~al.}{2023}]{xiao2023graph}
\begin{botherref}
\oauthor{\bsnm{Xiao}, \binits{S.}},
\oauthor{\bsnm{Lin}, \binits{H.}},
\oauthor{\bsnm{Wang}, \binits{C.}},
\oauthor{\bsnm{Wang}, \binits{S.}},
\oauthor{\bsnm{Rajapakse}, \binits{J.C.}}:
Graph neural networks with multiple prior knowledge for multi-omics data analysis.
IEEE Journal of Biomedical and Health Informatics
(2023)
\end{botherref}
\endbibitem

%%% 15
\bibitem[\protect\citeauthoryear{Wang et~al.}{2024}]{wang2024hypertmo}
\begin{barticle}
\bauthor{\bsnm{Wang}, \binits{H.}},
\bauthor{\bsnm{Lin}, \binits{K.}},
\bauthor{\bsnm{Zhang}, \binits{Q.}},
\bauthor{\bsnm{Shi}, \binits{J.}},
\bauthor{\bsnm{Song}, \binits{X.}},
\bauthor{\bsnm{Wu}, \binits{J.}},
\bauthor{\bsnm{Zhao}, \binits{C.}},
\bauthor{\bsnm{He}, \binits{K.}}:
\batitle{Hypertmo: a trusted multi-omics integration framework based on hypergraph convolutional network for patient classification}.
\bjtitle{Bioinformatics}
\bvolume{40}(\bissue{4}),
\bfpage{159}
(\byear{2024})
\end{barticle}
\endbibitem

%%% 16
\bibitem[\protect\citeauthoryear{Schlichtkrull et~al.}{2018}]{schlichtkrull2018modeling}
\begin{bchapter}
\bauthor{\bsnm{Schlichtkrull}, \binits{M.}},
\bauthor{\bsnm{Kipf}, \binits{T.N.}},
\bauthor{\bsnm{Bloem}, \binits{P.}},
\bauthor{\bsnm{Van Den~Berg}, \binits{R.}},
\bauthor{\bsnm{Titov}, \binits{I.}},
\bauthor{\bsnm{Welling}, \binits{M.}}:
\bctitle{Modeling relational data with graph convolutional networks}.
In: \bbtitle{The Semantic Web: 15th International Conference, ESWC 2018, Heraklion, Crete, Greece, June 3--7, 2018, Proceedings 15},
pp. \bfpage{593}--\blpage{607}
(\byear{2018}).
\bcomment{Springer}
\end{bchapter}
\endbibitem

%%% 17
\bibitem[\protect\citeauthoryear{Wang et~al.}{2019}]{wang2019heterogeneous}
\begin{bchapter}
\bauthor{\bsnm{Wang}, \binits{X.}},
\bauthor{\bsnm{Ji}, \binits{H.}},
\bauthor{\bsnm{Shi}, \binits{C.}},
\bauthor{\bsnm{Wang}, \binits{B.}},
\bauthor{\bsnm{Ye}, \binits{Y.}},
\bauthor{\bsnm{Cui}, \binits{P.}},
\bauthor{\bsnm{Yu}, \binits{P.S.}}:
\bctitle{Heterogeneous graph attention network}.
In: \bbtitle{The World Wide Web Conference},
pp. \bfpage{2022}--\blpage{2032}
(\byear{2019})
\end{bchapter}
\endbibitem

%%% 18
\bibitem[\protect\citeauthoryear{Sun et~al.}{2011}]{sun2011pathsim}
\begin{barticle}
\bauthor{\bsnm{Sun}, \binits{Y.}},
\bauthor{\bsnm{Han}, \binits{J.}},
\bauthor{\bsnm{Yan}, \binits{X.}},
\bauthor{\bsnm{Yu}, \binits{P.S.}},
\bauthor{\bsnm{Wu}, \binits{T.}}:
\batitle{Pathsim: Meta path-based top-k similarity search in heterogeneous information networks}.
\bjtitle{Proceedings of the VLDB Endowment}
\bvolume{4}(\bissue{11}),
\bfpage{992}--\blpage{1003}
(\byear{2011})
\end{barticle}
\endbibitem

%%% 19
\bibitem[\protect\citeauthoryear{Hu et~al.}{2020}]{hu2020heterogeneous}
\begin{bchapter}
\bauthor{\bsnm{Hu}, \binits{Z.}},
\bauthor{\bsnm{Dong}, \binits{Y.}},
\bauthor{\bsnm{Wang}, \binits{K.}},
\bauthor{\bsnm{Sun}, \binits{Y.}}:
\bctitle{Heterogeneous graph transformer}.
In: \bbtitle{Proceedings of the Web Conference 2020},
pp. \bfpage{2704}--\blpage{2710}
(\byear{2020})
\end{bchapter}
\endbibitem

%%% 20
\bibitem[\protect\citeauthoryear{Lv et~al.}{2021}]{lv2021we}
\begin{bchapter}
\bauthor{\bsnm{Lv}, \binits{Q.}},
\bauthor{\bsnm{Ding}, \binits{M.}},
\bauthor{\bsnm{Liu}, \binits{Q.}},
\bauthor{\bsnm{Chen}, \binits{Y.}},
\bauthor{\bsnm{Feng}, \binits{W.}},
\bauthor{\bsnm{He}, \binits{S.}},
\bauthor{\bsnm{Zhou}, \binits{C.}},
\bauthor{\bsnm{Jiang}, \binits{J.}},
\bauthor{\bsnm{Dong}, \binits{Y.}},
\bauthor{\bsnm{Tang}, \binits{J.}}:
\bctitle{Are we really making much progress? revisiting, benchmarking and refining heterogeneous graph neural networks}.
In: \bbtitle{Proceedings of the 27th ACM SIGKDD Conference on Knowledge Discovery \& Data Mining},
pp. \bfpage{1150}--\blpage{1160}
(\byear{2021})
\end{bchapter}
\endbibitem

%%% 21
\bibitem[\protect\citeauthoryear{Doshi-Velez and Kim}{2017}]{doshi2017towards}
\begin{botherref}
\oauthor{\bsnm{Doshi-Velez}, \binits{F.}},
\oauthor{\bsnm{Kim}, \binits{B.}}:
Towards a rigorous science of interpretable machine learning.
arXiv preprint arXiv:1702.08608
(2017)
\end{botherref}
\endbibitem

%%% 22
\bibitem[\protect\citeauthoryear{Colaprico et~al.}{2016}]{colaprico2016tcgabiolinks}
\begin{barticle}
\bauthor{\bsnm{Colaprico}, \binits{A.}},
\bauthor{\bsnm{Silva}, \binits{T.C.}},
\bauthor{\bsnm{Olsen}, \binits{C.}},
\bauthor{\bsnm{Garofano}, \binits{L.}},
\bauthor{\bsnm{Cava}, \binits{C.}},
\bauthor{\bsnm{Garolini}, \binits{D.}},
\bauthor{\bsnm{Sabedot}, \binits{T.S.}},
\bauthor{\bsnm{Malta}, \binits{T.M.}},
\bauthor{\bsnm{Pagnotta}, \binits{S.M.}},
\bauthor{\bsnm{Castiglioni}, \binits{I.}}, \betal:
\batitle{Tcgabiolinks: an r/bioconductor package for integrative analysis of tcga data}.
\bjtitle{Nucleic acids research}
\bvolume{44}(\bissue{8}),
\bfpage{71}--\blpage{71}
(\byear{2016})
\end{barticle}
\endbibitem

%%% 23
\bibitem[\protect\citeauthoryear{Hodes and Buckholtz}{2016}]{hodes2016accelerating}
\begin{barticle}
\bauthor{\bsnm{Hodes}, \binits{R.J.}},
\bauthor{\bsnm{Buckholtz}, \binits{N.}}:
\batitle{Accelerating medicines partnership: Alzheimer’s disease (amp-ad) knowledge portal aids alzheimer’s drug discovery through open data sharing}.
\bjtitle{Expert opinion on therapeutic targets}
\bvolume{20}(\bissue{4}),
\bfpage{389}--\blpage{391}
(\byear{2016})
\end{barticle}
\endbibitem

%%% 24
\bibitem[\protect\citeauthoryear{Petersen et~al.}{2010}]{petersen2010alzheimer}
\begin{barticle}
\bauthor{\bsnm{Petersen}, \binits{R.C.}},
\bauthor{\bsnm{Aisen}, \binits{P.S.}},
\bauthor{\bsnm{Beckett}, \binits{L.A.}},
\bauthor{\bsnm{Donohue}, \binits{M.C.}},
\bauthor{\bsnm{Gamst}, \binits{A.C.}},
\bauthor{\bsnm{Harvey}, \binits{D.J.}},
\bauthor{\bsnm{Jack~Jr}, \binits{C.}},
\bauthor{\bsnm{Jagust}, \binits{W.J.}},
\bauthor{\bsnm{Shaw}, \binits{L.M.}},
\bauthor{\bsnm{Toga}, \binits{A.W.}}, \betal:
\batitle{Alzheimer's disease neuroimaging initiative (adni) clinical characterization}.
\bjtitle{Neurology}
\bvolume{74}(\bissue{3}),
\bfpage{201}--\blpage{209}
(\byear{2010})
\end{barticle}
\endbibitem

%%% 25
\bibitem[\protect\citeauthoryear{Parker et~al.}{2009}]{parker2009supervised}
\begin{barticle}
\bauthor{\bsnm{Parker}, \binits{J.S.}},
\bauthor{\bsnm{Mullins}, \binits{M.}},
\bauthor{\bsnm{Cheang}, \binits{M.C.}},
\bauthor{\bsnm{Leung}, \binits{S.}},
\bauthor{\bsnm{Voduc}, \binits{D.}},
\bauthor{\bsnm{Vickery}, \binits{T.}},
\bauthor{\bsnm{Davies}, \binits{S.}},
\bauthor{\bsnm{Fauron}, \binits{C.}},
\bauthor{\bsnm{He}, \binits{X.}},
\bauthor{\bsnm{Hu}, \binits{Z.}}, \betal:
\batitle{Supervised risk predictor of breast cancer based on intrinsic subtypes}.
\bjtitle{Journal of clinical oncology}
\bvolume{27}(\bissue{8}),
\bfpage{1160}
(\byear{2009})
\end{barticle}
\endbibitem

%%% 26
\bibitem[\protect\citeauthoryear{Deng et~al.}{2017}]{deng2017firebrowser}
\begin{barticle}
\bauthor{\bsnm{Deng}, \binits{M.}},
\bauthor{\bsnm{Br{\"a}gelmann}, \binits{J.}},
\bauthor{\bsnm{Kryukov}, \binits{I.}},
\bauthor{\bsnm{Saraiva-Agostinho}, \binits{N.}},
\bauthor{\bsnm{Perner}, \binits{S.}}:
\batitle{Firebrowser: an r client to the broad institute’s firehose pipeline}.
\bjtitle{Database}
\bvolume{2017},
\bfpage{160}
(\byear{2017})
\end{barticle}
\endbibitem

%%% 27
\bibitem[\protect\citeauthoryear{Verhaak et~al.}{2010}]{verhaak2010integrated}
\begin{barticle}
\bauthor{\bsnm{Verhaak}, \binits{R.G.}},
\bauthor{\bsnm{Hoadley}, \binits{K.A.}},
\bauthor{\bsnm{Purdom}, \binits{E.}},
\bauthor{\bsnm{Wang}, \binits{V.}},
\bauthor{\bsnm{Qi}, \binits{Y.}},
\bauthor{\bsnm{Wilkerson}, \binits{M.D.}},
\bauthor{\bsnm{Miller}, \binits{C.R.}},
\bauthor{\bsnm{Ding}, \binits{L.}},
\bauthor{\bsnm{Golub}, \binits{T.}},
\bauthor{\bsnm{Mesirov}, \binits{J.P.}}, \betal:
\batitle{Integrated genomic analysis identifies clinically relevant subtypes of glioblastoma characterized by abnormalities in pdgfra, idh1, egfr, and nf1}.
\bjtitle{Cancer cell}
\bvolume{17}(\bissue{1}),
\bfpage{98}--\blpage{110}
(\byear{2010})
\end{barticle}
\endbibitem

%%% 28
\bibitem[\protect\citeauthoryear{Hares et~al.}{2017}]{hares2017overexpression}
\begin{barticle}
\bauthor{\bsnm{Hares}, \binits{K.}},
\bauthor{\bsnm{Miners}, \binits{J.S.}},
\bauthor{\bsnm{Cook}, \binits{A.J.}},
\bauthor{\bsnm{Rice}, \binits{C.}},
\bauthor{\bsnm{Scolding}, \binits{N.}},
\bauthor{\bsnm{Love}, \binits{S.}},
\bauthor{\bsnm{Wilkins}, \binits{A.}}:
\batitle{Overexpression of kinesin superfamily motor proteins in alzheimer’s disease}.
\bjtitle{Journal of Alzheimer's Disease}
\bvolume{60}(\bissue{4}),
\bfpage{1511}--\blpage{1524}
(\byear{2017})
\end{barticle}
\endbibitem

%%% 29
\bibitem[\protect\citeauthoryear{Hares et~al.}{2019}]{hares2019kif5a}
\begin{barticle}
\bauthor{\bsnm{Hares}, \binits{K.}},
\bauthor{\bsnm{Miners}, \binits{S.}},
\bauthor{\bsnm{Scolding}, \binits{N.}},
\bauthor{\bsnm{Love}, \binits{S.}},
\bauthor{\bsnm{Wilkins}, \binits{A.}}:
\batitle{Kif5a and klc1 expression in alzheimer’s disease: relationship and genetic influences}.
\bjtitle{AMRC Open Research}
\bvolume{1},
\bfpage{1}
(\byear{2019})
\end{barticle}
\endbibitem

%%% 30
\bibitem[\protect\citeauthoryear{Rocchio et~al.}{2019}]{rocchio2019gene}
\begin{barticle}
\bauthor{\bsnm{Rocchio}, \binits{F.}},
\bauthor{\bsnm{Tapella}, \binits{L.}},
\bauthor{\bsnm{Manfredi}, \binits{M.}},
\bauthor{\bsnm{Chisari}, \binits{M.}},
\bauthor{\bsnm{Ronco}, \binits{F.}},
\bauthor{\bsnm{Ruffinatti}, \binits{F.A.}},
\bauthor{\bsnm{Conte}, \binits{E.}},
\bauthor{\bsnm{Canonico}, \binits{P.L.}},
\bauthor{\bsnm{Sortino}, \binits{M.A.}},
\bauthor{\bsnm{Grilli}, \binits{M.}}, \betal:
\batitle{Gene expression, proteome and calcium signaling alterations in immortalized hippocampal astrocytes from an alzheimer’s disease mouse model}.
\bjtitle{Cell Death \& Disease}
\bvolume{10}(\bissue{1}),
\bfpage{24}
(\byear{2019})
\end{barticle}
\endbibitem

%%% 31
\bibitem[\protect\citeauthoryear{Liu et~al.}{2023}]{liu2023early}
\begin{barticle}
\bauthor{\bsnm{Liu}, \binits{Y.}},
\bauthor{\bsnm{Bilen}, \binits{M.}},
\bauthor{\bsnm{McNicoll}, \binits{M.-M.}},
\bauthor{\bsnm{Harris}, \binits{R.A.}},
\bauthor{\bsnm{Fong}, \binits{B.C.}},
\bauthor{\bsnm{Iqbal}, \binits{M.A.}},
\bauthor{\bsnm{Paul}, \binits{S.}},
\bauthor{\bsnm{Mayne}, \binits{J.}},
\bauthor{\bsnm{Walker}, \binits{K.}},
\bauthor{\bsnm{Wang}, \binits{J.}}, \betal:
\batitle{Early postnatal defects in neurogenesis in the 3xtg mouse model of alzheimer’s disease}.
\bjtitle{Cell Death \& Disease}
\bvolume{14}(\bissue{2}),
\bfpage{138}
(\byear{2023})
\end{barticle}
\endbibitem

%%% 32
\bibitem[\protect\citeauthoryear{Hermkens et~al.}{2019}]{hermkens2019profiling}
\begin{barticle}
\bauthor{\bsnm{Hermkens}, \binits{D.M.}},
\bauthor{\bsnm{Stam}, \binits{O.C.}},
\bauthor{\bsnm{Wit}, \binits{N.M.}},
\bauthor{\bsnm{Fontijn}, \binits{R.D.}},
\bauthor{\bsnm{Jongejan}, \binits{A.}},
\bauthor{\bsnm{Moerland}, \binits{P.D.}},
\bauthor{\bsnm{Mackaaij}, \binits{C.}},
\bauthor{\bsnm{Waas}, \binits{I.S.}},
\bauthor{\bsnm{Daemen}, \binits{M.J.}},
\bauthor{\bsnm{Vries}, \binits{H.E.}}:
\batitle{Profiling the unique protective properties of intracranial arterial endothelial cells}.
\bjtitle{Acta neuropathologica communications}
\bvolume{7},
\bfpage{1}--\blpage{16}
(\byear{2019})
\end{barticle}
\endbibitem

%%% 33
\bibitem[\protect\citeauthoryear{Li and De~Muynck}{2021}]{li2021differentially}
\begin{barticle}
\bauthor{\bsnm{Li}, \binits{Q.S.}},
\bauthor{\bsnm{De~Muynck}, \binits{L.}}:
\batitle{Differentially expressed genes in alzheimer’s disease highlighting the roles of microglia genes including olr1 and astrocyte gene cdk2ap1}.
\bjtitle{Brain, Behavior, \& Immunity-Health}
\bvolume{13},
\bfpage{100227}
(\byear{2021})
\end{barticle}
\endbibitem

%%% 34
\bibitem[\protect\citeauthoryear{Li et~al.}{2004}]{li2004analysis}
\begin{barticle}
\bauthor{\bsnm{Li}, \binits{H.}},
\bauthor{\bsnm{Wood}, \binits{C.L.}},
\bauthor{\bsnm{Getchell}, \binits{T.V.}},
\bauthor{\bsnm{Getchell}, \binits{M.L.}},
\bauthor{\bsnm{Stromberg}, \binits{A.J.}}:
\batitle{Analysis of oligonucleotide array experiments with repeated measures using mixed models}.
\bjtitle{BMC bioinformatics}
\bvolume{5},
\bfpage{1}--\blpage{12}
(\byear{2004})
\end{barticle}
\endbibitem

%%% 35
\bibitem[\protect\citeauthoryear{Sala~Frigerio et~al.}{2013}]{sala2013reduced}
\begin{barticle}
\bauthor{\bsnm{Sala~Frigerio}, \binits{C.}},
\bauthor{\bsnm{Lau}, \binits{P.}},
\bauthor{\bsnm{Salta}, \binits{E.}},
\bauthor{\bsnm{Tournoy}, \binits{J.}},
\bauthor{\bsnm{Bossers}, \binits{K.}},
\bauthor{\bsnm{Vandenberghe}, \binits{R.}},
\bauthor{\bsnm{Wallin}, \binits{A.}},
\bauthor{\bsnm{Bjerke}, \binits{M.}},
\bauthor{\bsnm{Zetterberg}, \binits{H.}},
\bauthor{\bsnm{Blennow}, \binits{K.}}, \betal:
\batitle{Reduced expression of hsa-mir-27a-3p in csf of patients with alzheimer disease}.
\bjtitle{Neurology}
\bvolume{81}(\bissue{24}),
\bfpage{2103}--\blpage{2106}
(\byear{2013})
\end{barticle}
\endbibitem

%%% 36
\bibitem[\protect\citeauthoryear{Wang et~al.}{2022}]{wang2022plasma}
\begin{barticle}
\bauthor{\bsnm{Wang}, \binits{L.}},
\bauthor{\bsnm{Zhen}, \binits{H.}},
\bauthor{\bsnm{Sun}, \binits{Y.}},
\bauthor{\bsnm{Rong}, \binits{S.}},
\bauthor{\bsnm{Li}, \binits{B.}},
\bauthor{\bsnm{Song}, \binits{Z.}},
\bauthor{\bsnm{Liu}, \binits{Z.}},
\bauthor{\bsnm{Li}, \binits{Z.}},
\bauthor{\bsnm{Ding}, \binits{J.}},
\bauthor{\bsnm{Yang}, \binits{H.}}, \betal:
\batitle{Plasma exo-mirnas correlated with ad-related factors of chinese individuals involved in a$\beta$ accumulation and cognition decline}.
\bjtitle{Molecular Neurobiology}
\bvolume{59}(\bissue{11}),
\bfpage{6790}--\blpage{6804}
(\byear{2022})
\end{barticle}
\endbibitem

%%% 37
\bibitem[\protect\citeauthoryear{Harati et~al.}{2022}]{harati2022mir}
\begin{barticle}
\bauthor{\bsnm{Harati}, \binits{R.}},
\bauthor{\bsnm{Hammad}, \binits{S.}},
\bauthor{\bsnm{Tlili}, \binits{A.}},
\bauthor{\bsnm{Mahfood}, \binits{M.}},
\bauthor{\bsnm{Mabondzo}, \binits{A.}},
\bauthor{\bsnm{Hamoudi}, \binits{R.}}:
\batitle{mir-27a-3p regulates expression of intercellular junctions at the brain endothelium and controls the endothelial barrier permeability}.
\bjtitle{PLoS One}
\bvolume{17}(\bissue{1}),
\bfpage{0262152}
(\byear{2022})
\end{barticle}
\endbibitem

%%% 38
\bibitem[\protect\citeauthoryear{Swarbrick et~al.}{2019}]{swarbrick2019systematic}
\begin{barticle}
\bauthor{\bsnm{Swarbrick}, \binits{S.}},
\bauthor{\bsnm{Wragg}, \binits{N.}},
\bauthor{\bsnm{Ghosh}, \binits{S.}},
\bauthor{\bsnm{Stolzing}, \binits{A.}}:
\batitle{Systematic review of mirna as biomarkers in alzheimer’s disease}.
\bjtitle{Molecular neurobiology}
\bvolume{56},
\bfpage{6156}--\blpage{6167}
(\byear{2019})
\end{barticle}
\endbibitem

%%% 39
\bibitem[\protect\citeauthoryear{Herrera-Espejo et~al.}{2019}]{herrera2019systematic}
\begin{barticle}
\bauthor{\bsnm{Herrera-Espejo}, \binits{S.}},
\bauthor{\bsnm{Santos-Zorrozua}, \binits{B.}},
\bauthor{\bsnm{{\'A}lvarez-Gonz{\'a}lez}, \binits{P.}},
\bauthor{\bsnm{Lopez-Lopez}, \binits{E.}},
\bauthor{\bsnm{Garcia-Orad}, \binits{{\'A}.}}:
\batitle{A systematic review of microrna expression as biomarker of late-onset alzheimer’s disease}.
\bjtitle{Molecular Neurobiology}
\bvolume{56},
\bfpage{8376}--\blpage{8391}
(\byear{2019})
\end{barticle}
\endbibitem

%%% 40
\bibitem[\protect\citeauthoryear{Kumar et~al.}{2013}]{kumar2013circulating}
\begin{barticle}
\bauthor{\bsnm{Kumar}, \binits{P.}},
\bauthor{\bsnm{Dezso}, \binits{Z.}},
\bauthor{\bsnm{MacKenzie}, \binits{C.}},
\bauthor{\bsnm{Oestreicher}, \binits{J.}},
\bauthor{\bsnm{Agoulnik}, \binits{S.}},
\bauthor{\bsnm{Byrne}, \binits{M.}},
\bauthor{\bsnm{Bernier}, \binits{F.}},
\bauthor{\bsnm{Yanagimachi}, \binits{M.}},
\bauthor{\bsnm{Aoshima}, \binits{K.}},
\bauthor{\bsnm{Oda}, \binits{Y.}}:
\batitle{Circulating mirna biomarkers for alzheimer's disease}.
\bjtitle{PloS one}
\bvolume{8}(\bissue{7}),
\bfpage{69807}
(\byear{2013})
\end{barticle}
\endbibitem

%%% 41
\bibitem[\protect\citeauthoryear{Gattuso et~al.}{2022}]{gattuso2022chronic}
\begin{barticle}
\bauthor{\bsnm{Gattuso}, \binits{G.}},
\bauthor{\bsnm{Falzone}, \binits{L.}},
\bauthor{\bsnm{Costa}, \binits{C.}},
\bauthor{\bsnm{Giamb{\`o}}, \binits{F.}},
\bauthor{\bsnm{Teodoro}, \binits{M.}},
\bauthor{\bsnm{Vivarelli}, \binits{S.}},
\bauthor{\bsnm{Libra}, \binits{M.}},
\bauthor{\bsnm{Fenga}, \binits{C.}}:
\batitle{Chronic pesticide exposure in farm workers is associated with the epigenetic modulation of hsa-mir-199a-5p}.
\bjtitle{International Journal of Environmental Research and Public Health}
\bvolume{19}(\bissue{12}),
\bfpage{7018}
(\byear{2022})
\end{barticle}
\endbibitem

%%% 42
\bibitem[\protect\citeauthoryear{Hewel et~al.}{2019}]{hewel2019common}
\begin{barticle}
\bauthor{\bsnm{Hewel}, \binits{C.}},
\bauthor{\bsnm{Kaiser}, \binits{J.}},
\bauthor{\bsnm{Wierczeiko}, \binits{A.}},
\bauthor{\bsnm{Linke}, \binits{J.}},
\bauthor{\bsnm{Reinhardt}, \binits{C.}},
\bauthor{\bsnm{Endres}, \binits{K.}},
\bauthor{\bsnm{Gerber}, \binits{S.}}:
\batitle{Common mirna patterns of alzheimer’s disease and parkinson’s disease and their putative impact on commensal gut microbiota}.
\bjtitle{Frontiers in neuroscience}
\bvolume{13},
\bfpage{113}
(\byear{2019})
\end{barticle}
\endbibitem

%%% 43
\bibitem[\protect\citeauthoryear{Giambra et~al.}{2023}]{giambra2023peritumoral}
\begin{barticle}
\bauthor{\bsnm{Giambra}, \binits{M.}},
\bauthor{\bsnm{Di~Cristofori}, \binits{A.}},
\bauthor{\bsnm{Valtorta}, \binits{S.}},
\bauthor{\bsnm{Manfrellotti}, \binits{R.}},
\bauthor{\bsnm{Bigiogera}, \binits{V.}},
\bauthor{\bsnm{Basso}, \binits{G.}},
\bauthor{\bsnm{Moresco}, \binits{R.M.}},
\bauthor{\bsnm{Giussani}, \binits{C.}},
\bauthor{\bsnm{Bentivegna}, \binits{A.}}:
\batitle{The peritumoral brain zone in glioblastoma: where we are and where we are going}.
\bjtitle{Journal of Neuroscience Research}
\bvolume{101}(\bissue{2}),
\bfpage{199}--\blpage{216}
(\byear{2023})
\end{barticle}
\endbibitem

%%% 44
\bibitem[\protect\citeauthoryear{Nimbalkar et~al.}{2021}]{nimbalkar2021differential}
\begin{barticle}
\bauthor{\bsnm{Nimbalkar}, \binits{V.P.}},
\bauthor{\bsnm{Kruthika}, \binits{B.S.}},
\bauthor{\bsnm{Sravya}, \binits{P.}},
\bauthor{\bsnm{Rao}, \binits{S.}},
\bauthor{\bsnm{Sugur}, \binits{H.S.}},
\bauthor{\bsnm{Verma}, \binits{B.K.}},
\bauthor{\bsnm{Chickabasaviah}, \binits{Y.T.}},
\bauthor{\bsnm{Arivazhagan}, \binits{A.}},
\bauthor{\bsnm{Kondaiah}, \binits{P.}},
\bauthor{\bsnm{Santosh}, \binits{V.}}:
\batitle{Differential gene expression in peritumoral brain zone of glioblastoma: role of serpina3 in promoting invasion, stemness and radioresistance of glioma cells and association with poor patient prognosis and recurrence}.
\bjtitle{Journal of Neuro-Oncology}
\bvolume{152},
\bfpage{55}--\blpage{65}
(\byear{2021})
\end{barticle}
\endbibitem

%%% 45
\bibitem[\protect\citeauthoryear{Papadimitriou and Kanellopoulou}{2023}]{papadimitriou2023protein}
\begin{barticle}
\bauthor{\bsnm{Papadimitriou}, \binits{E.}},
\bauthor{\bsnm{Kanellopoulou}, \binits{V.K.}}:
\batitle{Protein tyrosine phosphatase receptor zeta 1 as a potential target in cancer therapy and diagnosis}.
\bjtitle{International Journal of Molecular Sciences}
\bvolume{24}(\bissue{9}),
\bfpage{8093}
(\byear{2023})
\end{barticle}
\endbibitem

%%% 46
\bibitem[\protect\citeauthoryear{Xia et~al.}{2019}]{xia2019expression}
\begin{barticle}
\bauthor{\bsnm{Xia}, \binits{Z.}},
\bauthor{\bsnm{Ouyang}, \binits{D.}},
\bauthor{\bsnm{Li}, \binits{Q.}},
\bauthor{\bsnm{Li}, \binits{M.}},
\bauthor{\bsnm{Zou}, \binits{Q.}},
\bauthor{\bsnm{Li}, \binits{L.}},
\bauthor{\bsnm{Yi}, \binits{W.}},
\bauthor{\bsnm{Zhou}, \binits{E.}}:
\batitle{The expression, functions, interactions and prognostic values of ptprz1: a review and bioinformatic analysis}.
\bjtitle{Journal of Cancer}
\bvolume{10}(\bissue{7}),
\bfpage{1663}
(\byear{2019})
\end{barticle}
\endbibitem

%%% 47
\bibitem[\protect\citeauthoryear{Zeng et~al.}{2017}]{zeng2017tumour}
\begin{barticle}
\bauthor{\bsnm{Zeng}, \binits{A.}},
\bauthor{\bsnm{Yan}, \binits{W.}},
\bauthor{\bsnm{Liu}, \binits{Y.}},
\bauthor{\bsnm{Wang}, \binits{Z.}},
\bauthor{\bsnm{Hu}, \binits{Q.}},
\bauthor{\bsnm{Nie}, \binits{E.}},
\bauthor{\bsnm{Zhou}, \binits{X.}},
\bauthor{\bsnm{Li}, \binits{R.}},
\bauthor{\bsnm{Wang}, \binits{X.}},
\bauthor{\bsnm{Jiang}, \binits{T.}}, \betal:
\batitle{Tumour exosomes from cells harbouring ptprz1--met fusion contribute to a malignant phenotype and temozolomide chemoresistance in glioblastoma}.
\bjtitle{Oncogene}
\bvolume{36}(\bissue{38}),
\bfpage{5369}--\blpage{5381}
(\byear{2017})
\end{barticle}
\endbibitem

%%% 48
\bibitem[\protect\citeauthoryear{Cheng et~al.}{2022}]{cheng2022cysteine}
\begin{barticle}
\bauthor{\bsnm{Cheng}, \binits{X.}},
\bauthor{\bsnm{Ren}, \binits{Z.}},
\bauthor{\bsnm{Liu}, \binits{Z.}},
\bauthor{\bsnm{Sun}, \binits{X.}},
\bauthor{\bsnm{Qian}, \binits{R.}},
\bauthor{\bsnm{Cao}, \binits{C.}},
\bauthor{\bsnm{Liu}, \binits{B.}},
\bauthor{\bsnm{Wang}, \binits{J.}},
\bauthor{\bsnm{Wang}, \binits{H.}},
\bauthor{\bsnm{Guo}, \binits{Y.}}, \betal:
\batitle{Cysteine cathepsin c: a novel potential biomarker for the diagnosis and prognosis of glioma}.
\bjtitle{Cancer Cell International}
\bvolume{22}(\bissue{1}),
\bfpage{53}
(\byear{2022})
\end{barticle}
\endbibitem

%%% 49
\bibitem[\protect\citeauthoryear{Konduri et~al.}{2002}]{konduri2002modulation}
\begin{barticle}
\bauthor{\bsnm{Konduri}, \binits{S.D.}},
\bauthor{\bsnm{Yanamandra}, \binits{N.}},
\bauthor{\bsnm{Siddique}, \binits{K.}},
\bauthor{\bsnm{Joseph}, \binits{A.}},
\bauthor{\bsnm{Dinh}, \binits{D.H.}},
\bauthor{\bsnm{Olivero}, \binits{W.C.}},
\bauthor{\bsnm{Gujrati}, \binits{M.}},
\bauthor{\bsnm{Kouraklis}, \binits{G.}},
\bauthor{\bsnm{Swaroop}, \binits{A.}},
\bauthor{\bsnm{Kyritsis}, \binits{A.P.}}, \betal:
\batitle{Modulation of cystatin c expression impairs the invasive and tumorigenic potential of human glioblastoma cells}.
\bjtitle{Oncogene}
\bvolume{21}(\bissue{57}),
\bfpage{8705}--\blpage{8712}
(\byear{2002})
\end{barticle}
\endbibitem

%%% 50
\bibitem[\protect\citeauthoryear{Li et~al.}{2016}]{li2016retracted}
\begin{barticle}
\bauthor{\bsnm{Li}, \binits{Y.}},
\bauthor{\bsnm{Zhang}, \binits{X.}},
\bauthor{\bsnm{Chen}, \binits{D.}},
\bauthor{\bsnm{Ma}, \binits{C.}}:
\batitle{Retracted article: Let-7a suppresses glioma cell proliferation and invasion through tgf-$\beta$/smad3 signaling pathway by targeting hmga2}.
\bjtitle{Tumor Biology}
\bvolume{37}(\bissue{6}),
\bfpage{8107}--\blpage{8119}
(\byear{2016})
\end{barticle}
\endbibitem

%%% 51
\bibitem[\protect\citeauthoryear{Lee et~al.}{2011}]{lee2011let}
\begin{barticle}
\bauthor{\bsnm{Lee}, \binits{S.-T.}},
\bauthor{\bsnm{Chu}, \binits{K.}},
\bauthor{\bsnm{Oh}, \binits{H.-J.}},
\bauthor{\bsnm{Im}, \binits{W.-S.}},
\bauthor{\bsnm{Lim}, \binits{J.-Y.}},
\bauthor{\bsnm{Kim}, \binits{S.-K.}},
\bauthor{\bsnm{Park}, \binits{C.-K.}},
\bauthor{\bsnm{Jung}, \binits{K.-H.}},
\bauthor{\bsnm{Lee}, \binits{S.K.}},
\bauthor{\bsnm{Kim}, \binits{M.}}, \betal:
\batitle{Let-7 microrna inhibits the proliferation of human glioblastoma cells}.
\bjtitle{Journal of neuro-oncology}
\bvolume{102},
\bfpage{19}--\blpage{24}
(\byear{2011})
\end{barticle}
\endbibitem

%%% 52
\bibitem[\protect\citeauthoryear{Xi et~al.}{2019}]{xi2019joint}
\begin{barticle}
\bauthor{\bsnm{Xi}, \binits{X.}},
\bauthor{\bsnm{Chu}, \binits{Y.}},
\bauthor{\bsnm{Liu}, \binits{N.}},
\bauthor{\bsnm{Wang}, \binits{Q.}},
\bauthor{\bsnm{Yin}, \binits{Z.}},
\bauthor{\bsnm{Lu}, \binits{Y.}},
\bauthor{\bsnm{Chen}, \binits{Y.}}:
\batitle{Joint bioinformatics analysis of underlying potential functions of hsa-let-7b-5p and core genes in human glioma}.
\bjtitle{Journal of translational medicine}
\bvolume{17},
\bfpage{1}--\blpage{16}
(\byear{2019})
\end{barticle}
\endbibitem

%%% 53
\bibitem[\protect\citeauthoryear{Wang et~al.}{2020}]{wang2020emerging}
\begin{botherref}
\oauthor{\bsnm{Wang}, \binits{Y.}},
\oauthor{\bsnm{Zeng}, \binits{G.}},
\oauthor{\bsnm{Jiang}, \binits{Y.}}:
The emerging roles of mir-125b in cancers.
Cancer management and research,
1079--1088
(2020)
\end{botherref}
\endbibitem

%%% 54
\bibitem[\protect\citeauthoryear{Akers et~al.}{2013}]{akers2013mir}
\begin{barticle}
\bauthor{\bsnm{Akers}, \binits{J.C.}},
\bauthor{\bsnm{Ramakrishnan}, \binits{V.}},
\bauthor{\bsnm{Kim}, \binits{R.}},
\bauthor{\bsnm{Skog}, \binits{J.}},
\bauthor{\bsnm{Nakano}, \binits{I.}},
\bauthor{\bsnm{Pingle}, \binits{S.}},
\bauthor{\bsnm{Kalinina}, \binits{J.}},
\bauthor{\bsnm{Hua}, \binits{W.}},
\bauthor{\bsnm{Kesari}, \binits{S.}},
\bauthor{\bsnm{Mao}, \binits{Y.}}, \betal:
\batitle{Mir-21 in the extracellular vesicles (evs) of cerebrospinal fluid (csf): a platform for glioblastoma biomarker development}.
\bjtitle{PloS one}
\bvolume{8}(\bissue{10}),
\bfpage{78115}
(\byear{2013})
\end{barticle}
\endbibitem

%%% 55
\bibitem[\protect\citeauthoryear{Ben-Hamo et~al.}{2016}]{ben2016hsa}
\begin{barticle}
\bauthor{\bsnm{Ben-Hamo}, \binits{R.}},
\bauthor{\bsnm{Zilberberg}, \binits{A.}},
\bauthor{\bsnm{Cohen}, \binits{H.}},
\bauthor{\bsnm{Efroni}, \binits{S.}}:
\batitle{hsa-mir-9 controls the mobility behavior of glioblastoma cells via regulation of mapk14 signaling elements}.
\bjtitle{Oncotarget}
\bvolume{7}(\bissue{17}),
\bfpage{23170}
(\byear{2016})
\end{barticle}
\endbibitem

%%% 56
\bibitem[\protect\citeauthoryear{Kipf and Welling}{2016}]{kipf2016semi}
\begin{botherref}
\oauthor{\bsnm{Kipf}, \binits{T.N.}},
\oauthor{\bsnm{Welling}, \binits{M.}}:
Semi-supervised classification with graph convolutional networks.
arXiv preprint arXiv:1609.02907
(2016)
\end{botherref}
\endbibitem

%%% 57
\bibitem[\protect\citeauthoryear{Sohn}{2016}]{sohn2016improved}
\begin{botherref}
\oauthor{\bsnm{Sohn}, \binits{K.}}:
Improved deep metric learning with multi-class n-pair loss objective.
Advances in neural information processing systems
\textbf{29}
(2016)
\end{botherref}
\endbibitem

%%% 58
\bibitem[\protect\citeauthoryear{Veli{\v{c}}kovi{\'c} et~al.}{2017}]{velivckovic2017graph}
\begin{botherref}
\oauthor{\bsnm{Veli{\v{c}}kovi{\'c}}, \binits{P.}},
\oauthor{\bsnm{Cucurull}, \binits{G.}},
\oauthor{\bsnm{Casanova}, \binits{A.}},
\oauthor{\bsnm{Romero}, \binits{A.}},
\oauthor{\bsnm{Lio}, \binits{P.}},
\oauthor{\bsnm{Bengio}, \binits{Y.}}:
Graph attention networks.
arXiv preprint arXiv:1710.10903
(2017)
\end{botherref}
\endbibitem

%%% 59
\bibitem[\protect\citeauthoryear{Kingma and Ba}{2014}]{kingma2014adam}
\begin{botherref}
\oauthor{\bsnm{Kingma}, \binits{D.P.}},
\oauthor{\bsnm{Ba}, \binits{J.}}:
Adam: A method for stochastic optimization.
arXiv preprint arXiv:1412.6980
(2014)
\end{botherref}
\endbibitem

\end{thebibliography}

\end{document}